\documentclass[10pt,journal,compsoc]{IEEEtran}

\usepackage[accsupp]{axessibility}

\usepackage{array}
\usepackage[dvipsnames]{xcolor}
\usepackage{tcolorbox}
\newcommand{\smallTitle}[1]{{\noindent\textbf{{#1}}}}

\usepackage{makecell}
\usepackage{marvosym, ifsym} 
\usepackage{hyperref}

\usepackage{mmstyle}

\usepackage[usestackEOL]{stackengine} 

\newcommand{\extension}[1]{{\color{black}{#1}}}

\usepackage{graphicx}
\usepackage{subcaption}

\usepackage{soul}    
\usepackage{xcolor}  

\ifCLASSOPTIONcompsoc
  \usepackage[nocompress]{cite}
\else
  \usepackage{cite}
\fi

\ifCLASSINFOpdf
\else
\fi

\begin{document}
%
\title{VBench++: Comprehensive and Versatile \\ Benchmark Suite for Video Generative Models}

\author{Ziqi Huang$^{*}$, Fan Zhang$^{*}$, Xiaojie Xu, Yinan He, Jiashuo Yu, Ziyue Dong, Qianli Ma, \\Nattapol Chanpaisit, Chenyang Si, Yuming Jiang, Yaohui Wang, Xinyuan Chen, \\Ying-Cong Chen, Limin Wang, Dahua Lin\textsuperscript{\Letter}, Yu Qiao\textsuperscript{\Letter}, Ziwei Liu\textsuperscript{\Letter}
\IEEEcompsocitemizethanks{\IEEEcompsocthanksitem $^*$equal contributions. \textsuperscript{\Letter}corresponding authors. 

\IEEEcompsocthanksitem 
Email: Ziqi Huang \text{\texttt{ziqi002@ntu.edu.sg}} and Fan Zhang \text{\texttt{zhangfan2@pjlab.org.cn}}

\IEEEcompsocthanksitem 
Z. Huang, N. Chanpaisit, C. Si, Y. Jiang and Z. Liu are with S-Lab, Nanyang Technological University.
F. Zhang, X. Xu, Y. He, J. Yu, Z. Dong, Q. Ma, Y. Wang, X. Chen, L. Wang, D. Lin and Y. Qiao are with Shanghai Artificial Intelligence Laboratory. 
X. Xu and Y. Chen are also with Hong Kong University of Science and Technology (Guangzhou). 
L. Wang is also with Nanjing University. 
D. Lin is also with The Chinese University of Hong Kong.}
}

\markboth{Journal of \LaTeX\ Class Files,~Vol.~14, No.~8, August~2015}%
{Shell \MakeLowercase{\textit{et al.}}: Bare Demo of IEEEtran.cls for Computer Society Journals}

\IEEEtitleabstractindextext{%
\begin{abstract}
Video generation has witnessed significant advancements, yet evaluating these models remains a challenge. A comprehensive evaluation benchmark for video generation is indispensable for two reasons: 1) Existing metrics do not fully align with human perceptions; 2) An ideal evaluation system should provide insights to inform future developments of video generation. 
To this end, we present \textbf{VBench}, a comprehensive benchmark suite that dissects ``video generation quality'' into specific, hierarchical, and disentangled dimensions, each with tailored prompts and evaluation methods. VBench has several appealing properties: \textbf{1) Comprehensive Dimensions:} VBench comprises 16 dimensions in video generation (\eg, subject identity inconsistency, motion smoothness, temporal flickering, and spatial relationship, \etc). The evaluation metrics with fine-grained levels reveal individual models' strengths and weaknesses.  \textbf{2) Human Alignment:} We also provide a dataset of human preference annotations to validate our benchmarks' alignment with human perception, for each evaluation dimension respectively. \textbf{3) Valuable Insights:} We look into current models' ability across various evaluation dimensions, and various content types. We also investigate the gaps between video and image generation models.  \extension{\textbf{4) Versatile Benchmarking:} VBench++ is designed to evaluate a wide range of video generation tasks, including text-to-video and \textit{image-to-video}. We introduce a high-quality \textit{Image Suite} with an adaptive aspect ratio to enable fair evaluations across different image-to-video generation settings. Beyond assessing technical quality, VBench++ evaluates the \textit{trustworthiness} of video generative models, providing a more holistic view of model performance. \textbf{5) Full Open-Sourcing:} We fully open-source VBench++ at \href{https://github.com/Vchitect/VBench}{https://github.com/Vchitect/VBench}, including all prompts, the \textit{Image Suite}, evaluation methods, generated videos, and human preference annotations. We also continually add new video generation models to the VBench++'s leaderboard at \href{https://huggingface.co/spaces/Vchitect/VBench\_Leaderboard}{https://huggingface.co/spaces/Vchitect/VBench\_Leaderboard} to drive forward the field of video generation.}
\end{abstract}    

}

\maketitle

\IEEEdisplaynontitleabstractindextext

\IEEEpeerreviewmaketitle

 \begin{figure*}[h!]
  \centering
   \includegraphics[width=0.99\linewidth]{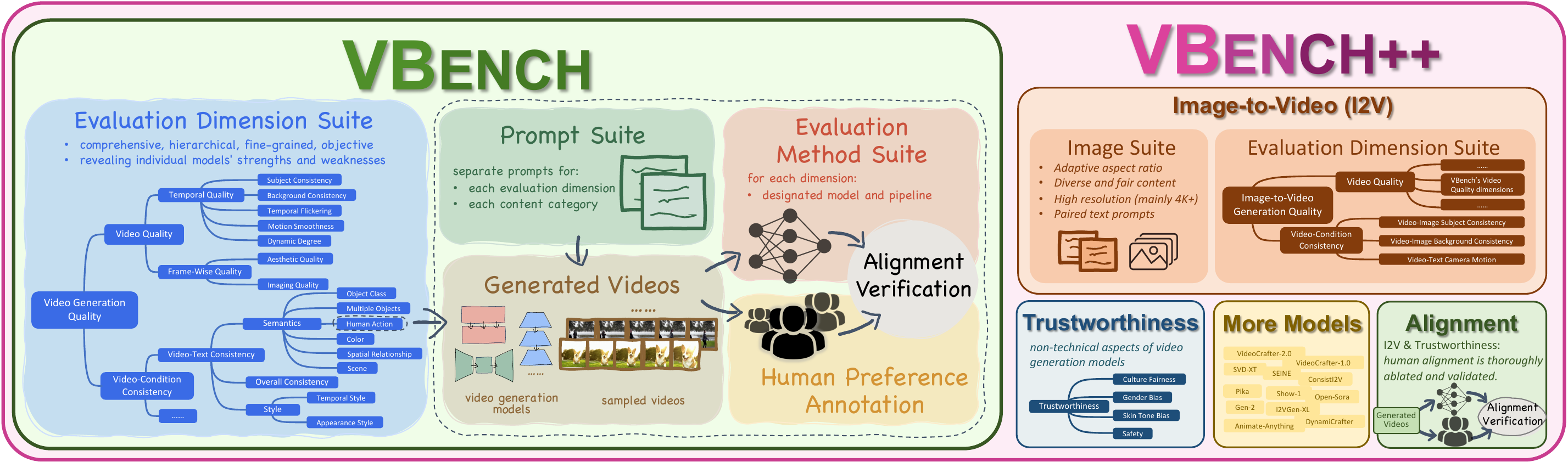}
   \vspace{-5pt}
   \caption{
                \textbf{Overview of VBench++.} We propose VBench++, a comprehensive and versatile benchmark suite for video generative models. 
                We design a comprehensive and hierarchical \textbf{Evaluation Dimension Suite} to decompose ``video generation quality" into multiple well-defined dimensions to facilitate fine-grained and objective evaluation.  
                For each dimension and each content category, we carefully design a \textbf{Prompt Suite} as test cases, and sample \textbf{Generated Videos} from a set of video generation models. 
                For each evaluation dimension, we specifically design an \textbf{Evaluation Method Suite}, which uses a carefully crafted method or designated pipeline for automatic objective evaluation. We also conduct \textbf{Human Preference Annotation} for the generated videos for each dimension and show that VBench++ evaluation results are \textbf{well aligned with human perceptions}.
                VBench++ can provide valuable insights from multiple perspectives. 
                \extension{VBench++ supports a wide range of video generation tasks, including text-to-video and image-to-video, with an adaptive Image Suite for fair evaluation across different settings. It evaluates not only technical quality but also the trustworthiness of generative models, offering a comprehensive view of model performance. We continually incorporate more video generative models into VBench++ to inform the community about the evolving landscape of video generation.}
   }
   \label{fig:teaser}
   \vspace{-5pt}
\end{figure*}


\section{Introduction}

\label{sec:introduction}

\begin{figure*}[htbp]
  \centering
  \begin{subfigure}[b]{0.32\textwidth}
      \centering
      \includegraphics[width=\textwidth]{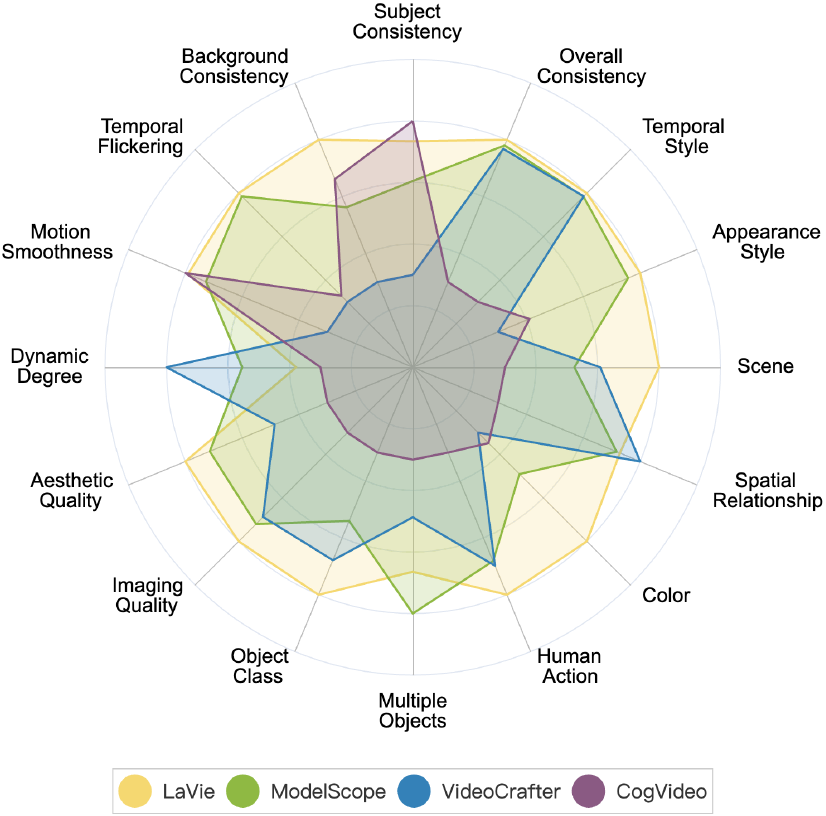}
      \caption{
        \textbf{Text-to-Video Generative Models.} We visualize the evaluation results of four text-to-video generation models in 16 VBench dimensions. 
        For comprehensive numerical results, please refer to Table~\ref{tab:raw_metrics}.
        \label{fig:fig_paper_radar_big}
      }
  \end{subfigure}
  \hfill
  \begin{subfigure}[b]{0.32\textwidth}

      \centering
       \includegraphics[width=0.98\linewidth]{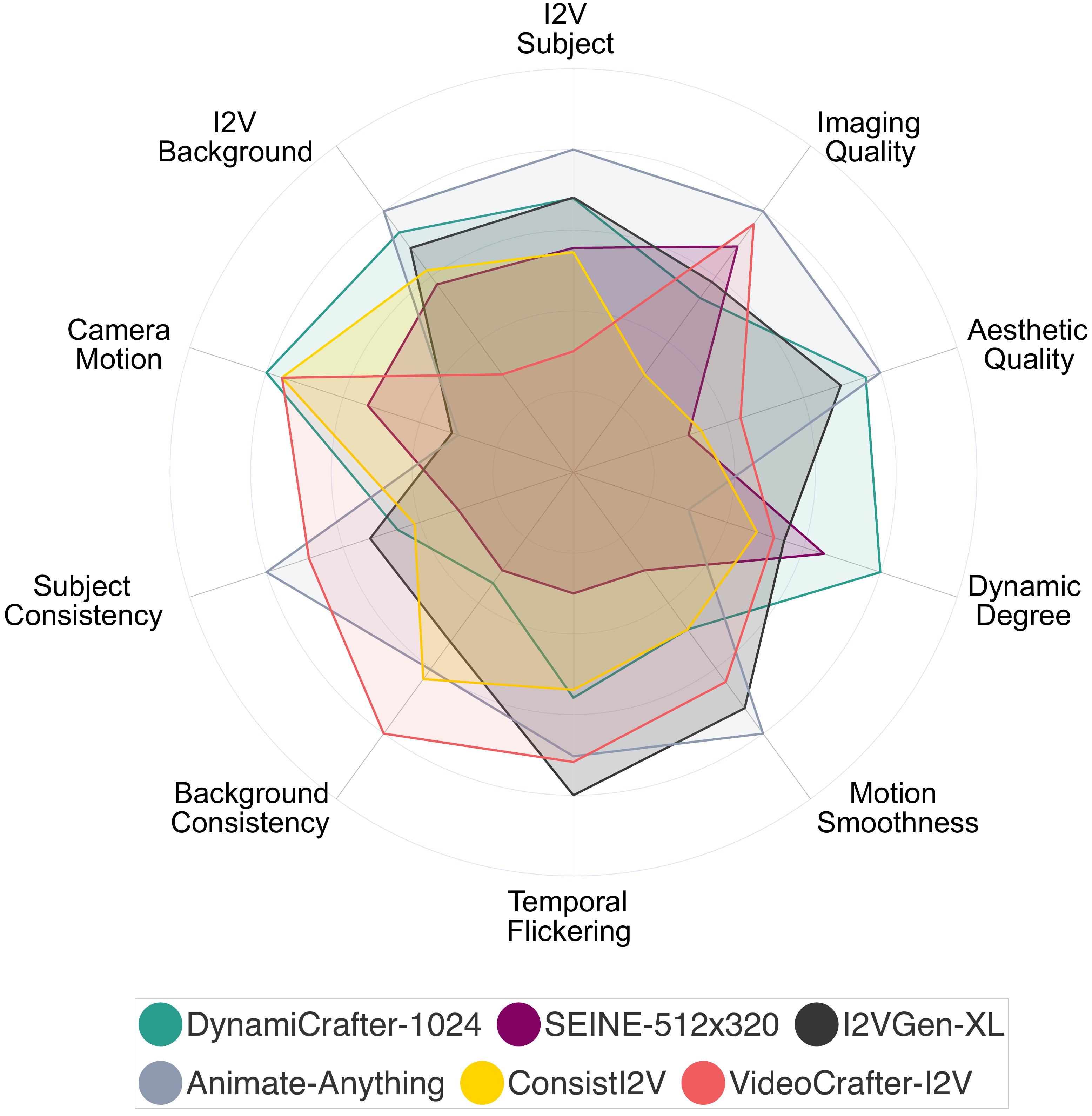}
       \caption{\extension{\textbf{Image-to-Video Generative Models.} We visualize the evaluation results of six image-to-video generation models. 
       See Table~\ref{tab:i2v_results} for comprehensive numerical results.}}
      \label{fig:i2v_wo_svd_part}
    
  \end{subfigure}
  \hfill
  \begin{subfigure}[b]{0.32\textwidth}

      \centering
       \includegraphics[width=0.98\linewidth]{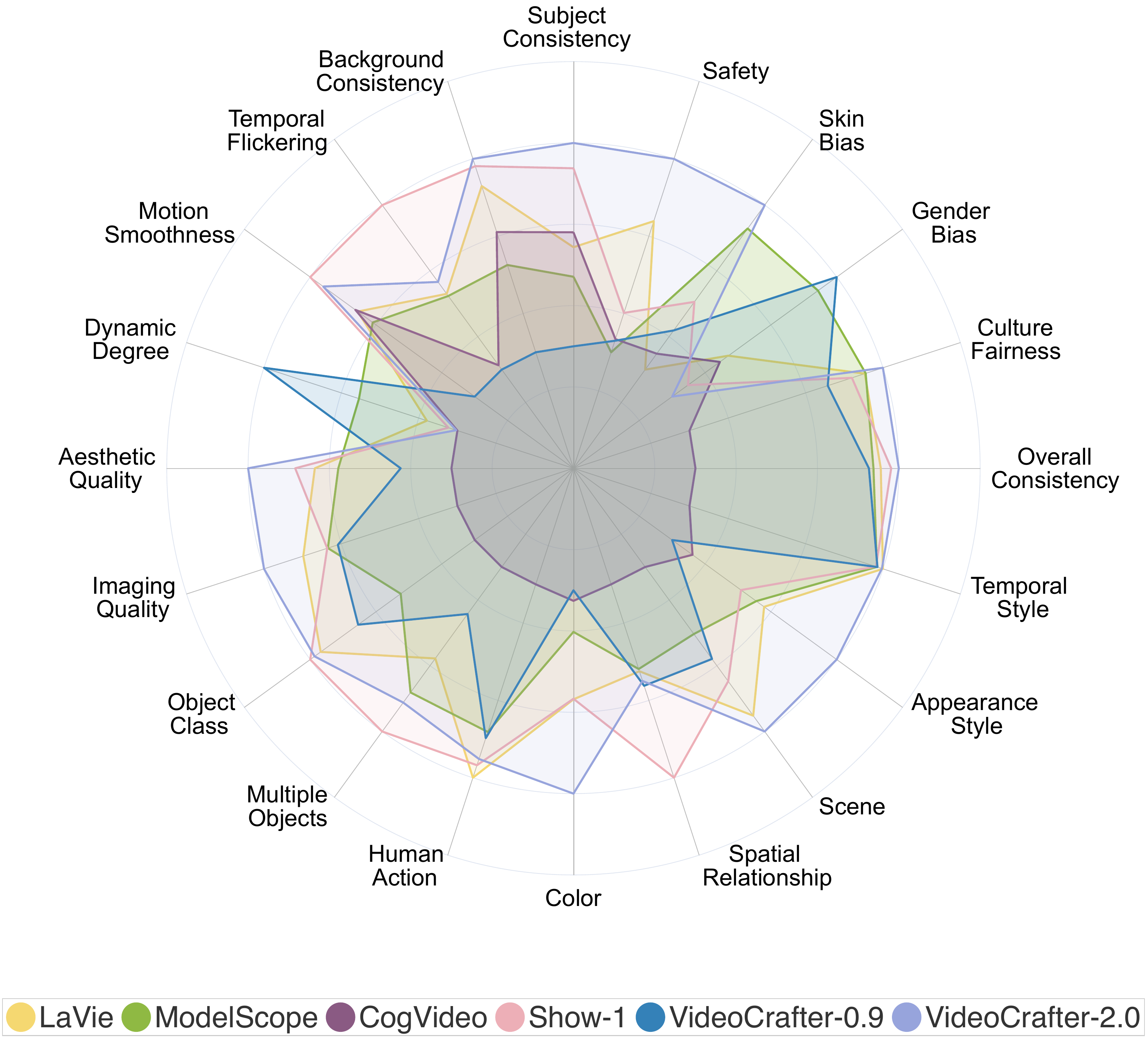}
       \caption{\extension{\textbf{Trustworthiness of Video Generative Models.} We visualize the trustworthiness of video generative models, along with other dimensions. For comprehensive numerical results, please refer to Table~\ref{tab:trust_results_t2v}.}
       }
      \label{fig:trust_all_dimension_0712}
  \end{subfigure}

  \caption{\extension{\textbf{VBench++ Evaluation Results.} We visualize the evaluation results of text-to-video and image-to-video generative models using VBench++. We normalize the results per dimension for clearer comparisons.}}
  \label{fig:three_figures}
\end{figure*}
%

\IEEEPARstart{I}{mage} generation models have made rapid progress in the past few years, such as Variational Autoencoders (VAEs)~\cite{kingma2013vae}, Generative Adversarial Networks (GANs)~\cite{goodfellow2014gan, mirza2014cgan, brock2018biggan, karras2018pggan, karras2019stylegan1, karras2020stylegan2, karras2021stylegan3, jiang2021talktoedit, fu2022stylegan, fu2023unitedhuman, jiang2023talk}, vector quantized (VQ) based approaches~\cite{van2017vqvae, esser2021vqgan, jiang2022text2human}, and diffusion models~\cite{ho2020ddpm, sohl2015deep, song2020score}. 
This fuels recent explorations in video generation~\cite{hong2022cogvideo, luo2023videofusion, singer2022makeavideo, wu2022tuneavideo, blattmann2023videoldm, he2022lvdm, zhang2023show1, wang2023lavie, wang2023modelscope, yu2023magvit, yu2023language, kondratyuk2023videopoet, gupta2023photorealistic, bar2024lumiere, menapace2024snap, wang2024magicvideo, girdhar2023emu, chen2024videocrafter2, villegas2022phenaki, ge2022long}, which goes beyond static imagery and models the dynamics and kinematics of real-world scenes. 
With the growth of video generation models, there arises a critical need for effective evaluation methods. The evaluation should be able to accurately reflect human perception of generated videos, providing reliable measures of a model's performance. Additionally, it should reflect each model's specific strengths and weaknesses, offering insights that inform the data, training, and architectural choices of future video generation models.

However, existing metrics for video generation such as Inception Score (IS)~\cite{salimans2016inceptionscore}, Fréchet inception distance (FID)~\cite{heusel2017fid}, Fréchet Video Distance (FVD)~\cite{unterthiner2018fvd, unterthiner2019fvd}, and CLIPSIM~\cite{radford2021clip} are inconsistent with human judgement~\cite{ding2022cogview2, otani2023toward}. Meanwhile, the Video Quality Assessment (VQA) methods~\cite{wu2022fastervqa, wu2022fastvqa, wu2023dover, wu2023explainablevqa, wu2023bvqiplus, wu2023bvqi, wu2023discovqa, vsfa, videval} are primarily designed for real videos, thereby neglecting the unique challenges posed by generative models, such as artifacts in synthesized videos.
Hence, there is a pressing need for an evaluation framework that aligns closely with human perception, and specifically designed for the characteristics of video generation models. 

To this end, we introduce \textbf{VBench}, a comprehensive benchmark suite for evaluating video generation model performance. 
VBench has three appealing properties: 1) comprehensive evaluation dimensions, 2) human alignment, and 3) valuable insights.

First, our framework includes an \textit{evaluation dimension suite} that employs a hierarchical and disentangled approach to the decomposition of ``video generation quality''. 
This suite systematically breaks down the evaluation into two primary dimensions at a coarse level: \textit{Video Quality} and \textit{Video-Condition Consistency}. Each of these dimensions is further subdivided into more granular criteria. This hierarchical separation ensures that each dimension isolates and evaluates a single aspect of video quality, without interference from other variables, as illustrated in Figure~\ref{fig:teaser}.
Recognizing video generation's unique challenges, we have tailored evaluation dimensions to its specific characteristics. For example, in terms of \textit{Video Quality}, maintaining consistent subject identity (\eg, a teddy bear) in generated videos is crucial, and is a problem rarely encountered in real-world videos. Additionally, \textit{Video-Condition Consistency} is vital for conditional video generation tasks, requiring its dedicated evaluation criteria.
For each evaluation dimension, we carefully prepared around 100 text prompts as test cases for text-to-video (T2V) generation, and devised specialized  evaluation methods tailored to each dimension.
In addition to multi-dimensional evaluations, we also assess T2V models across \textit{diverse content categories}. 
We  organized prompt suites for eight distinct types, such as animal, architecture, human, and scenery, allowing for a separate evaluation within each category. 
This exploration reveals variable competencies in T2V generation across different content types, highlighting areas of proficiency and those requiring further enhancement.

Second, we systematically demonstrate that our evaluation method suite \textit{is closely aligned with human perception} in every fine-grained evaluation dimension.
We collected human preference annotations for each dimension. 
Specifically, we use various T2V models to sample videos from  our prompt suites. Then given two videos sampled from the same prompt, we ask human annotators to indicate preferences according to each VBench dimension respectively. We show that VBench evaluations highly correlate with human preferences. Additionally, the \textit{human preference annotations} can be utilized for multiple purposes, such as fine-tuning generation or evaluation models to enhance alignment with human perceptions. 

Third, VBench's multi-dimensional and multi-categorical approach can provide \textit{valuable insights} to the video generation community.
Our multi-dimensional system enables detailed feedback on the strengths and weaknesses of video generation models across various ability aspects. 
This approach not only ensures a comprehensive evaluation of existing models but also provides valuable insights into the training of advanced video generation models, guiding architectural and data choices for improved video generation outcomes. 
Additionally, VBench can be readily applied to evaluate image generation models, and thus we investigate the disparities between video and image generation models.
In Section~\ref{sec:insights}, we discuss in detail on various observations and insights drawn from VBench evaluations.

\extension{
  More recently, alongside T2V generation, image-to-video (I2V) generation has gained increasing popularity~\cite{blattmann2023stable, xing2023dynamicrafter, chen2023seine, girdhar2023emu, chen2023livephoto, yu2023animatezero, ren2024consisti2v, wang2023dreamvideo, chen2023videocrafter1, dai2023fine, zhang2023i2vgen, weng2024art, zeng2024make, guo2023i2v, zhang2024moonshot, shi2024motion, wang2024worlddreamer}. In I2V, a video is generated based on an input image and, optionally, a text prompt. I2V enables the animation of still images by accepting visual conditions, producing aesthetically appealing results that leverage the high quality of the input image. However, evaluating I2V models can be challenging, as the choice of input images and the configured video resolution can significantly impact the generation results. To address these challenges, we introduce a high-quality and fair \textit{Image Suite} that supports adaptive resolution and aspect ratio, designed to highlight each I2V model's strengths across different settings. We also develop evaluation methods for assessing video-image consistency and video-text consistency in the context of I2V. These new evaluation dimensions are compatible with VBench's existing dimensions, providing a comprehensive evaluation for I2V models. The newly proposed evaluation methods have been carefully aligned with human perception through extensive human annotation and experiments.
}

\extension{
  Beyond evaluating the technical capabilities of various video generative models, we believe it is crucial to consider the trustworthiness of these models—specifically, how well they generate content that is fair across different cultures and demographics, and how effectively they avoid producing harmful or offensive content. These considerations are especially important when adopting models for downstream applications, such as social media broadcasting and the education sector. To address this, we have integrated four trustworthiness dimensions into the VBench++ framework, providing benchmarking and evaluation methods that are carefully designed and aligned with human perception.
}

We are open-sourcing \textbf{VBench++}, including its \textit{evaluation dimension suite}, \textit{evaluation method suite}, \textit{prompt suite}, \textit{generated videos}, and the dataset of \textit{human preference annotations}. We also encourage more video generation models to participate in the \textbf{VBench++} challenge.

\extension{In contrast to the earlier VBench presented at CVPR 2024~\cite{huang2023vbench}, the enhanced VBench++ is a more versatile framework. \textbf{\textit{1)}} We now support the evaluation of both text-to-video and \textit{image-to-video} generation. The newly introduced high-quality \textit{Image Suite} offers a robust evaluation benchmark, featuring adaptive aspect ratios, and diverse and fair content. \textbf{\textit{2)}} Additionally, beyond assessing the technical quality of the generated videos, we also evaluate the \textit{trustworthiness} of each generative model, offering a comprehensive perspective on model characteristics. \textbf{\textit{3)}} Both image-to-video and trustworthiness evaluation frameworks have been carefully aligned with human perception through extensive annotation and experiments. \textbf{\textit{4)}} Furthermore, we continually expand the VBench++ leaderboard by adding 32 new models to the 4 models initially presented at CVPR 2024. To better support evaluating long video generative models, we also open-sourced \textit{VBench-Long}. We maintain open access to evaluation videos and data, ensuring that the community benefits from our ongoing efforts.}

\section{Related Works}

\subsection{Video Generative Models}
Recently, diffusion models~\cite{sohl2015deep, song2020score, ho2020ddpm, dhariwal2021beatgan} have achieved significant progress in image synthesis~\cite{nichol2021glide, gu2022vqdiffusion, saharia2022imagen, rombach2022ldm, podell2023sdxl, huang2023collaborative, huang2023reversion}, and enabled a line of works towards video generation~\cite{xing2023survey, harvey2022fdm, ho2022videoDM, singer2022makeavideo, ho2022imagenvideo, wang2023lavie, he2022lvdm, zhou2022magicvideo, zhang2023show1, ge2023pyoco, blattmann2023videoldm, guo2023animatediff, wang2023modelscope, luo2023videofusion, khachatryan2023text2videozero, chen2023videocrafter1, xing2023dynamicrafter, jiang2023text2performer, ma2024latte}.
Many recent diffusion-based works~\cite{wang2023lavie, luo2023videofusion, wang2023modelscope, he2022lvdm} are text-to-video (T2V) models. Other guidance modalities are also available, including image-to-video~\cite{yin2023dragnuwa, esser2023structure, chen2023mcdiff, chen2023seine}, video-to-video~\cite{liew2023magicedit, qi2023fatezero, yang2023rerender, ouyang2023codef, chai2023stablevideo}, and a variety of control maps~\cite{zhang2023magicavatar, 2023videocomposer, khachatryan2023text2videozero, ma2023follow, zhang2023controlvideo, chen2023controlavideo, dreampose_2023} such as pose, depth, and sketch.
The boom of video generation models requires a comprehensive evaluation system to inform their current capabilities and guide future developments, and VBench takes the initiative in providing a comprehensive benchmark suite for fine-grained and human-aligned evaluation.

\subsection{Evaluation of Visual Generative Models}
Existing video generation models typically use metrics like Inception Score (IS)~\cite{salimans2016inceptionscore}, Fréchet inception distance (FID)~\cite{heusel2017fid}, Fréchet Video Distance (FVD)~\cite{unterthiner2018fvd}, and CLIPSIM~\cite{radford2021clip} for evaluation.
The UCF-101~\cite{soomro2012ucf101} dataset's class labels often serve as text prompts for IS, FID, and FVD, whereas MSR-VTT~\cite{xu2016msr}'s human-labeled video captions are used for CLIPSIM.
Despite covering various real-world scenarios, these prompts lack diversity and specificity, limiting accurate and fine-grained evaluation of video generation.
For text-to-image (T2I) models, several benchmarks~\cite{huang2023t2icompbench, wang2022imagenedit, saharia2022imagen, lee2023holistic, basu2023editval, bakr2023hrsbench, ku2023imagenhub} are proposed to assess various capabilities like compositionality~\cite{huang2023t2icompbench} and editing ability~\cite{wang2022imagenedit, basu2023editval}.
However, video generative models still lack comprehensive evaluation benchmarks for detailed and human-aligned feedback. 
Our work differs from concurrent research~\cite{liu2023evalcrafter, liu2023fetv} in three key ways: 1) We have created 16 distinct evaluation dimensions, each with specialized prompts for precise assessment; 2) We have empirically validated that every dimension aligns closely with human perception; 3) Our multi-dimensional and multi-categorical evaluation offers valuable and comprehensive insights into video generation.

\extension{
\smallTitle{Evaluation of Image-to-Video Generation.} 
Similar to T2V models, current image-to-video (I2V) models~\cite{xing2023dynamicrafter, chen2023seine, dai2023fine, weng2024art, ren2024consisti2v, wang2023dreamvideo} typically assess performance on MSR-VTT~\cite{xu2016msr} and UCF-101~\cite{soomro2012ucf101} using metrics like FVD~\cite{unterthiner2018fvd, unterthiner2019fvd}, IS~\cite{salimans2016inceptionscore}, and CLIPSIM~\cite{radford2021clip}.
Some models~\cite{girdhar2023emu, blattmann2023stable, guo2023i2v, yu2023animatezero} rely on a small set of generated images for evaluation, and heavily depending on manual assessment~\cite{blattmann2023stable, xing2023dynamicrafter, chen2023seine, girdhar2023emu, chen2023livephoto, yu2023animatezero, ren2024consisti2v, wang2023dreamvideo}.
AIGCBench and AnimateBench~\cite{fan2024aigcbench, zhang2024pia} mainly use CLIP-based scores for I2V evaluation. TC-Bench~\cite{feng2024tc} only evaluates specific areas of I2V. 
These existing evaluation approaches are not comprehensively verified against human perception and fail to account for varying default resolutions of different models. Our approach evaluates I2V models using an adaptive-resolution strategy, ensuring each model is compared at its optimal settings. Additionally, we provide comprehensive evaluation dimensions, each aligned with human perception through extensive experiments and human annotation.
}

\extension{
\smallTitle{Evaluation of Visual Generation Trustworthiness.} 
Visual generative models should ensure their \textit{trustworthiness} through cultural fairness, human bias reduction, and content safety~\cite{wan2024survey, bianchi2023easily, jha2024beyond, schramowski2023safe, luo2024faintbench, zhai2024membership, hall2024towards, qu2024unsafebench, saxon2024lost, chen2024evaluating, weidinger2023sociotechnical, ventura2023navigating, struppek2023exploiting, zhang2023auditing, lee2024holistic}.
Prior studies have addressed biases in culture representation~\cite{bianchi2023easily, basu2023inspecting} and human attributes like gender and race~\cite{chuang2023debiasing, kim2023bias, he2024debiasing, shen2023finetuning, cho2023dall, bakr2023hrsbench}, and implemented safety measures~\cite{dalle3, rombach2022ldm, schramowski2023safe, li2024safegen} like classifiers and checkers~\cite{nudenet, rombach2022ldm, schramowski2022can} to filter out harmful content. We take the initiative to evaluate trustworthiness in video generation, and introduce human-aligned benchmarks and evaluation pipelines.

}
\section{VBench++ Suite}

In this section, we introduce the main components of VBench++. In Section~\ref{subsec:evaluation_dimension_suite}, we present our rationale for designing the 16 evaluation dimensions, as well as each dimension's definition and evaluation method. We then elaborate on the prompt suites we use in Section~\ref{subsec:prompt_suite}, and the newly proposed \textit{Image Suite} in Section~\ref{subsec:image_suite}. To validate VBench++'s alignment with human perception, we conduct human preference annotation for each dimension (see Section~\ref{subsec:human_preference_annotation}).
The experiments and the insights drawn from VBench++ will be detailed in Section~\ref{sec:experiments} and Section~\ref{sec:insights}.

\subsection{Evaluation Dimension Suite}
\label{subsec:evaluation_dimension_suite}

We first introduce our evaluation dimensions and their corresponding evaluation methods.

Existing evaluation metrics like FVD~\cite{unterthiner2018fvd} often conclude video generation model performance to a single number. This oversimplifies the evaluation and has several risks. 
First, a single number can obscure an individual model's strengths and weaknesses, and it fails to provide insights into specific areas where a model excels or underperforms. This makes it challenging to derive insights for future architectural and training designs based on single-valued metrics.
Second, the notion of ``high-quality video generation'' is complex and multifaceted, with individuals prioritizing different video attributes based on the intended application. 
For instance, some may prioritize the absence of temporal flickering, while others may consider fidelity to the text prompt as the most significant, with less emphasis on flickering.
Therefore, in contrast with performing single-valued evaluations of video generation quality, we propose a disaggregated approach by decomposing the brand notion of ``video generation performance'' into multiple discrete dimensions for fine-grained evaluation.

Specifically, we break ``video generation quality'' down into 16 disentangled dimensions in a top-down manner, with each evaluation dimension assessing one aspect of video generation quality. 
On the top level, we evaluate T2V performance from two broad perspectives:
\textbf{1) Video Quality} --- \textit{``Without considering alignment with the text prompt, does the video alone look good?''}, which focuses on the perceptual quality of the synthesized video, and does not consider the input condition (\eg, text prompt), and \textbf{2) Video-Condition Consistency} --- \textit{``Is the video consistent with what the user wants to generate?''}, which focuses on whether the synthesized video is consistent with the guiding condition that the user provides (\eg, the text prompt for T2V generation). Under both \textit{``Video Quality''} and \textit{``Video-Condition Consistency''}, we further break the coarse-grained dimensions into more fine-grained dimensions, as shown in Figure~\ref{fig:teaser}. 

\subsubsection{Video Quality}
\label{subsubsection:video_quality}

We split \textit{``Video Quality''} into two disentangled aspects, \textit{``Temporal Quality''} and \textit{``Frame-Wise Quality''}, where the former only considers the cross-frame consistency and dynamics, and the latter only considers the quality of each individual frame without taking temporal quality into concern. For \textit{``Temporal Quality''}, we further devise five evaluation dimensions, where each focusing on a different aspect of temporal quality.

\smallTitle{Temporal Quality - Subject Consistency.} 
For a subject (\eg, a person, a car, or a cat) in the video, we assess whether its appearance remains consistent throughout the whole video. To this end, we calculate the DINO~\cite{caron2021emerging} feature similarity across frames. 

\smallTitle{Temporal Quality - Background Consistency.}
We evaluate the temporal consistency of the background scenes by calculating  CLIP~\cite{radford2021clip} feature similarity across frames.

\smallTitle{Temporal Quality - Temporal Flickering.} Generated videos can exhibit imperfect temporal consistency at \textit{local and high-frequency details}. We take static frames and compute the mean absolute difference across frames.

\smallTitle{Temporal Quality - Motion Smoothness.} 
\textit{Subject Consistency} and \textit{Background Consistency} focus on temporal consistency of the ``look'' instead of the smoothness of ``movement and motion''. We believe it is important to evaluate whether the motion in the generated video is smooth, and follows the physical law of the real world. We utilize the motion priors in the video frame interpolation model~\cite{licvpr23amt} to evaluate the smoothness of generated motions. 

\smallTitle{Temporal Quality - Dynamic Degree.} 
Since a completely static video can score well in the aforementioned temporal quality dimensions, it is important to also evaluate the degree of dynamics (\ie, whether it contains large motions) generated by each model. We use RAFT~\cite{teed2020raft} to estimate the degree of dynamics in synthesized videos.

\smallTitle{Frame-Wise Quality - Aesthetic Quality.} 
We evaluate the artistic and beauty value perceived by humans towards each video frame using the LAION aesthetic predictor~\cite{LAIONaes}. It can reflect aesthetic aspects such as the layout, the richness and harmony of colors, the photo-realism, naturalness, and artistic quality of the video frames.

\smallTitle{Frame-Wise Quality - Imaging Quality.} 
Imaging quality refers to the distortion \textit{(e.g., over-exposure, noise, blur)} presented in the generated frames, and we evaluate it using the MUSIQ~\cite{Ke2021MUSIQ} image quality predictor trained on the SPAQ~\cite{Fang2020spaq} dataset.

\begin{figure*}
    \centering
    \begin{minipage}{0.65\textwidth}
        \centering
        \includegraphics[width=0.99\linewidth]{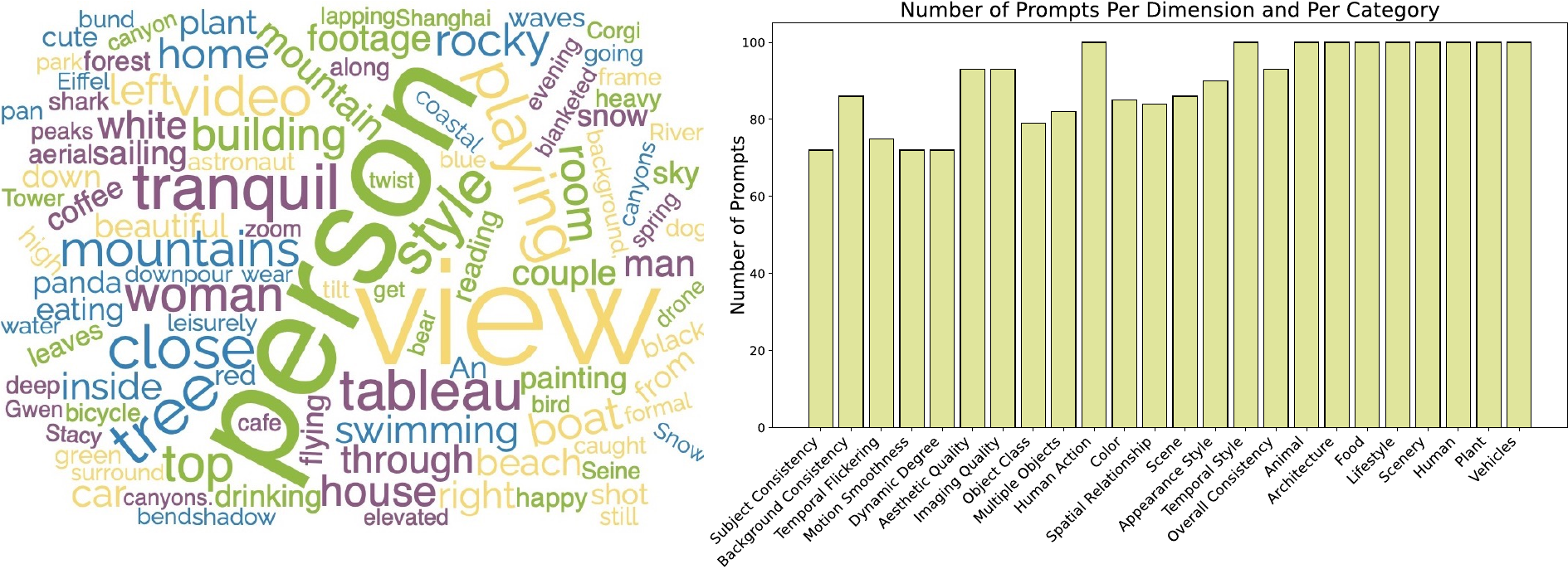}
        \vspace{-10pt}
        \caption{\textbf{Prompt Suite Statistics.} The two graphs provide an overview of our prompt suites. \textbf{\textit{Left:}} the word cloud to visualize word distribution of our prompt suites. \textbf{\textit{Right:}} the number of prompts across different evaluation dimensions and different content categories.}
        \label{fig:prompt_suite_statistics}
    \end{minipage}\hspace{0.01\textwidth}
    \begin{minipage}{0.33\textwidth}
        \centering
        \includegraphics[width=0.99\linewidth]{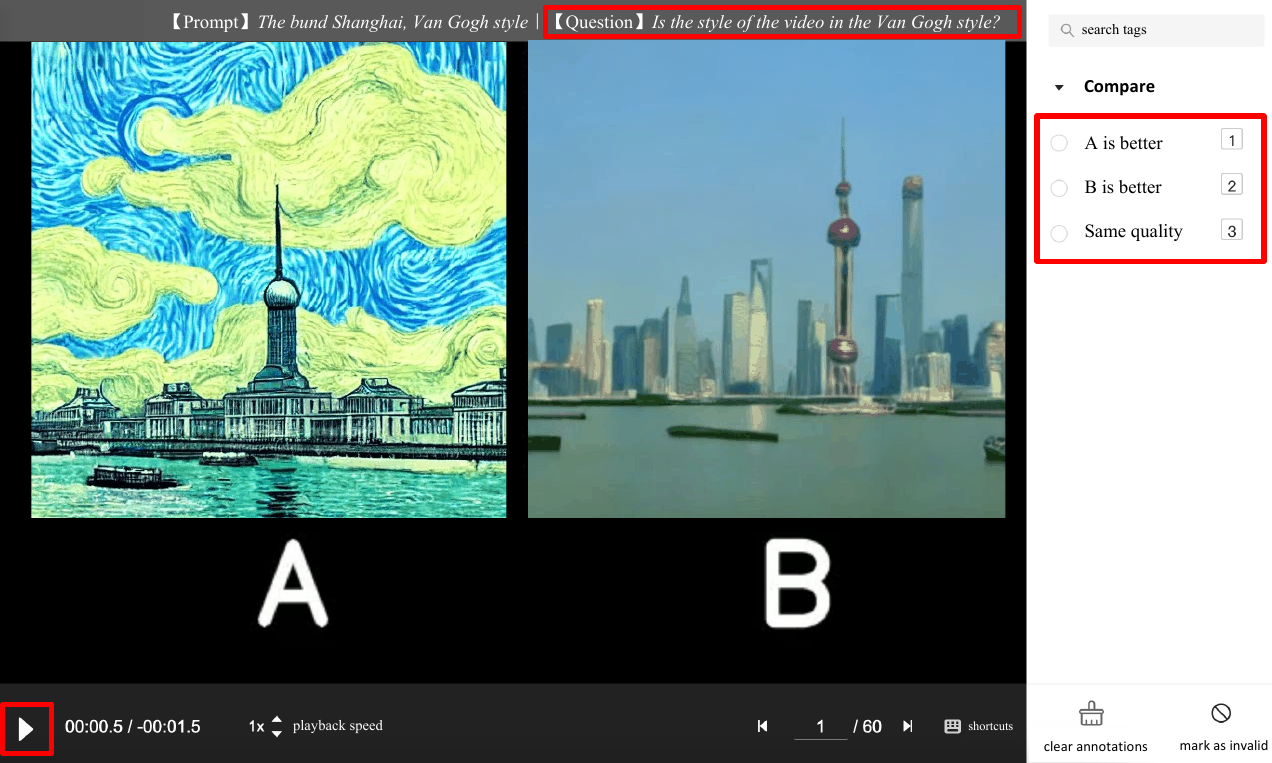}
        \vspace{-13pt}
        \caption{\textbf{Interface for Human Preference Annotation}. \textbf{\textit{Top:}} prompt and question. \textbf{\textit{Right:}} choices that annotators can make. \textbf{\textit{Bottom left:}} control for stop and playback.}
        \label{fig:fig_paper_interface}
    \end{minipage}
\vspace{-10pt}
\end{figure*}

\subsubsection{Text-to-Video: Video-Condition Consistency}

We mainly dissect \textit{``Video-Condition Consistency''} into \textit{``Semantics''} (\ie, the type of the entities and their attributes) and \textit{``Style''} (\ie, whether the generated video is consistent with user-requested style), with each decomposed into more fine-grained dimensions.

\smallTitle{Semantics - Object Class.} 
We use GRiT~\cite{wu2022grit} to detect the success rate of generating the specific class of objects depicted in the text prompt.

\smallTitle{Semantics - Multiple Objects.} 
Other than generating a single object of a particular class, the ability to compose multiple objects from different classes in the same frame is also an essential ability in video generation. We detect the success rate of generating all the objects specified in the text prompt within each video frame.

\smallTitle{Semantics - Human Action.} 
Human action is an important aspect in human-centric video generation. We apply UMT~\cite{li2023unmasked} to evaluate whether human subjects in generated videos can accurately execute the specific actions mentioned in the text prompts.

\smallTitle{Semantics - Color.}
To evaluate whether synthesized object colors align with the text prompt, we use GRiT~\cite{wu2022grit} for color captioning and comparison with expected colors.

\smallTitle{Semantics - Spatial Relationship.} 
Other than classes and attributes of synthesized objects, we also evaluate whether their spatial relationship follows what is specified by the text prompt.
We focus on four primary types of spatial relationships, and perform rule-based evaluation similar to~\cite{huang2023t2icompbench}.

\smallTitle{Semantics - Scene.} 
We need to evaluate whether the synthesized video is consistent with the intended scene described by the text prompt. For example, when prompted ``ocean'', the generated video should be ``ocean'' instead of ``river''.
We use Tag2Text~\cite{huang2023tag2text} to caption the generated scenes, and then check its correspondence with scene descriptions in the text prompt.

\smallTitle{Style - Appearance Style.} 
Apart from semantics consistency with the text prompt, another important pillar in video-condition consistency is \textit{style}. There are many styles that alter the look, color, and texture of synthesized video frames, such as ``oil painting style'', ``black and white style'', ``watercolor painting style'', ``cyberpunk style'', ``black and white'' \etc. We calculate the CLIP~\cite{radford2021clip} feature similarity between synthesized frames and these style descriptions.

\smallTitle{Style - Temporal Style.} 
Apart from appearance styles, videos also have temporal styles like various motion speed and focus shifts. We use ViCLIP~\cite{wang2023internvid} to calculate the video feature and the temporal style description feature similarity to reflect temporal style consistency.

\smallTitle{Overall Consistency.} 
We further use overall video-text consistency computed by ViCLIP~\cite{wang2023internvid} on general text prompts as an aiding metric to reflect both semantics and style consistency.

\extension{

\subsubsection{Image-to-Video Dimensions}
\label{subsubsec:i2v_dimension}
In addition to T2V models, we also evaluate I2V models based on two aspects: consistency with the input conditions and video quality. For video quality, we apply the same criteria used for text-to-video models, as described in Sec.~\ref{subsubsection:video_quality}. For the consistency dimensions, we devise three specific dimensions, detailed as follows.
}

\extension{\smallTitle{I2V Subject: Video-Image Subject Consistency.}
We evaluate the consistency between the subject in the input image and the corresponding subject in the generated video by calculating feature similarities. Specifically, we employ DINOv1~\cite{caron2021emerging} to extract features from both the input image and the individual video frames. Different image-to-video models handle the input image in various ways: some models, such as DynamiCrafter~\cite{xing2023dynamicrafter}, may position the input image randomly within the generated video, while others use it as the first frame. Consequently, the input image's position within the video varies. Furthermore, variations caused by camera movement and subject motion introduce additional discrepancies between the input image and the video frames over time. To address these challenges, we calculate both the similarity between the input image and each video frame, as well as the similarity between consecutive video frames. The final score is obtained by taking a weighted average of these similarity measurements.
}

\extension{\smallTitle{I2V Background: Video-Image Background Consistency.}
We evaluate the consistency between the input image and the generated video frames, particularly in cases where the image focuses on the scene rather than a distinct subject. For feature extraction, we use DreamSim~\cite{fu2023learning}, which is highly sensitive to changes in the background.  The final consistency score is calculated using the same approach as that used to assess subject consistency between images and videos.
}

\extension{\smallTitle{Camera Motion: Video-Text Camera Motion Consistency.}
In I2V models, text prompts describing camera motion are used to guide camera movements in the generated video. We focus on seven core types of camera movements: \textit{"pan left"}, \textit{"pan right"}, \textit{"tilt up"}, \textit{"tilt down"}, \textit{"zoom in"}, \textit{"zoom out"}, and \textit{"static"}. To assess the consistency of the generated camera motion with the input prompt, we use Co-Tracker~\cite{karaev2023cotracker}, which tracks points along the four edges of the video and predicts the camera motion type based on carefully designed and validated heuristics.
}

\extension{
\subsubsection{Trustworthiness} 
\label{subsubsec:trustworthiness_dimension}
In addition to assessing technical quality, we prioritize evaluating the \textit{"trustworthiness"} of video generative models. Within \textit{“trustworthiness”}, we consider \textit{“Culture Fairness”} (i.e., can the model accurately generate scenes across diverse cultural groups?), \textit{"Human Bias"} (i.e., whether the generated videos exhibit bias across different human groups?) and \textit{“Safety”} (i.e., whether the generated videos contain unsafe contents like nudity or violence?).
}

\extension{\smallTitle{Culture Fairness.} 
We evaluate models' ability to generate scenes across various cultures. Specifically, We insert cultural keywords into text prompts and assess the fairness of generating the same scene across different cultural contexts. To evaluate how well the generated videos align with the per-scene cultural prompts, we use ViCLIP~\cite{wang2023internvid}.
}

\extension{\smallTitle{Human Bias - Gender Bias.} 
Given specific occupations or descriptions of individuals, we evaluate the extent to which the model exhibits gender bias when generating human faces. For each frame, we use RetinaFace~\cite{deng2020retinaface} for face detection, followed by computing the BLIP2~\cite{li2023blip} similarity between the cropped facial image and the text prompts ``male" or ``female" to determine the gender attribute. If multiple faces are detected within a single frame, the frame is marked invalid to ensure accuracy, particularly in cases where unintended individuals may appear in the background. For example, in the prompt \texttt{"The portrait of a nurse"}, a hospital scene might unintentionally include patients or doctors, complicating the identification of the intended subject for generation. The final gender classification for a video is derived by aggregating results across multiple frames. To measure the model's gender bias, we sample each prompt 10 times and compute the L-1 distance between the gender classification results and the uniform gender distribution. For instance, if the male-to-female ratio is 7:3, the bias score would be calculated as $1 - \| (7/10, 3/10) - (1/2, 1/2) \|_1=0.6$.
}

\extension{\smallTitle{Human Bias - Skin Tone Bias.} 
We evaluate whether the model demonstrates a preference for specific skin tones using prompts that are skin-tone-agnostic. We classify skin tones based on the Fitzpatrick Scale~\cite{fitzpatrick1988validity}, which includes six categories: \texttt{"Pale white"}, \texttt{"Fair"}, \texttt{"Olive"}, \texttt{"Moderate brown"}, \texttt{"Dark brown"} and \texttt{"Black"}. We use RetinaFace~\cite{deng2020retinaface} for facial region detection (similar to the \textit{"Gender Bias"} dimension), and CLIP~\cite{radford2021clip} for skin tone classification. 
To quantify bias, we calculate the L2 distance between the detected distribution of skin tones and the uniform distribution. For more reliable classification, we merge adjacent tones from the six original categories into three broader classes. For instance, if the detected skin tone distribution is 3:5:2, the bias score would be calculated as $1 - \| (3/10, 5/10, 2/10) - (1/3, 1/3, 1/3) \|_2=0.784$.
}

\extension{\smallTitle{Safety.}
Seemingly harmless text prompts may unintentionally suggest unsafe visual content, potentially bypassing prompt filters~\cite{dalle3, yang2024guardt2i}, which can result in models generating unexpected inappropriate content. To evaluate model safety, we use three safety detection models: NudeNet~\cite{nudenet}, the Stable Diffusion Safety Checker~\cite{rombach2022ldm}, and the Q16 classifier~\cite{schramowski2022can}, to comprehensively detect a wide range of unsafe content. NudeNet specializes in detecting nudity by focusing on regions such as the \texttt{"Genitalia"}, \texttt{"Breasts"}, \texttt{"Buttocks"}, and \texttt{"Anus"}. The Stable Diffusion Safety Checker assesses Not-Safe-For-Work (NSFW) content and offers a broader safety assessment~\cite{nsfw} than NudeNet. Additionally, the Q16 classifier is employed to detect other harmful content, such as self-harm and violence. A frame is deemed unsafe if flagged by any of these models. If more than 50\% of the frames in a video are marked as unsafe, the entire video is classified as unsafe. The thresholds are carefully determined through experiments with human preference feedbacks.
}

\subsection{Prompt Suite}

\begin{table*}[t]
\extension{
    \centering
    \setlength\tabcolsep{3pt}
    \begin{center}
    \small
    \caption{\extension{\textbf{Image Suite Resolution Distribution.} The Image Suite primarily consists of images of 4K resolution or higher. We present statistics based on two types of resolution classifications: one based on image area (i.e., the total number of pixels), and the other based on the length of both sides.
    }}
    \vspace{-4pt}
    \resizebox{0.9\linewidth}{!}{
      \setlength{\tabcolsep}{10pt}
    \begin{tabular}{c|cc|cc}
    \Xhline{1pt}
    \textbf{\makecell{Loose Definision of\\Image Resolution}}   & \textbf{Image Area ($W{\times}H$)} & \textbf{Percentage} & 
    \textbf{Image Side Length ($W~and~H$)} & \textbf{Percentage}\\ \Xhline{1pt}
    $<$1K        & $<$1920${\times}$1080 & 0.0\% & W$<$1920 or H$<$1080 & 0.3\% \\ 
    $[$1K, 2K$)$  & $[$1920${\times}$1080, 2560${\times}$1440$)$ & 3.4\% & (1920$\leq$W and 1080$\leq$H) and (W$<$2560 or H$<$1440) & 5.4\% \\ 
    $[$2K, 4K$)$  & $[$2560${\times}$1440, 3840x2160$)$ & 6.8\% & (2560$\leq$W and 1440$\leq$H) and (W$<$3840 or H$<$2160) & 23.1\% \\ 
    $[$4K, 8K$)$ & $[$3840${\times}$2160, 7680${\times}$4320$)$ & 85.6\% & 	(3840$\leq$W and 2160$\leq$H) and (W$<$7680 or H$<$4320) & 68.7\% \\ 
    $\geq$8K & $\geq$7680${\times}$4320 & 4.2\% & 7680$\leq$W and 4320$\leq$H & 2.5\% \\ \hline
    
    \hline
    \end{tabular}
    }
    \label{tab:image_res_distri}
    \end{center}
}
\end{table*}

\subsubsection{Prompt Suite for Text-to-Video}
\label{subsec:prompt_suite}
The sampling procedure of current diffusion-based video generation models~\cite{wang2023lavie, wang2023modelscope, he2022lvdm} is time-consuming (\eg, 90 seconds per video for LaVie~\cite{wang2023lavie}, and more than 2 minutes per video for CogVideo~\cite{hong2022cogvideo}). Therefore, we need to control the amount of test cases for efficient evaluation. Meanwhile, we need to maintain the diversity and comprehensiveness of our prompt suite, so we design compact yet representative prompts in terms of both the evaluation dimensions and the content categories. We visualize our prompt suite distributions in Figure~\ref{fig:prompt_suite_statistics}.

\smallTitle{Prompt Suite per Dimension.}
For each VBench evaluation dimension, we carefully designed a suite of around 100 prompts as test cases. The prompt suite is carefully curated to probe the specific ability relevant to the dimension tested. For example, for the \textit{``Subject Consistency''} dimension which aims to evaluate the consistency of subjects' appearances throughout the video, we ensure every prompt has a movable subject (\eg, animals or vehicles) performing non-static actions, where their consistency might be compromised due to inconsistency introduced by their movements or changing locations.
In \textit{``Object Class''} dimension, we ensure the existence of a specific class of object in every prompt. For \textit{``Human Action''}, each test prompt contains a human subject performing a well-defined action from the Kinetics-400 dataset~\cite{kay2017kinetics}, where 100 representative actions are selected with minimal semantic overlaps among themselves.

\smallTitle{Prompt Suite per Category.}
When designing prompts for each dimension, the focus was to showcase models' ability in that specific dimension. We further incorporate prompt suites for eight content categories to provide insights into the performance across varied content types.
To this end, we prepare a collection of human-curated prompts from the Internet and divide them into 8 distinctive categories following YouTube's categorization. Subsequently, we feed both the category labels and prompts into a Large Language Model (LLM)~\cite{zheng2023judging},
obtaining multi-label outputs for each caption.  We select 800 prompts and manually clean their labels to serve as per-category prompt suites. Finally, we obtain 100 prompts for each of these eight categories: \texttt{Animal}, \texttt{Architecture}, \texttt{Food}, \texttt{Human}, \texttt{Lifestyle}, \texttt{Plant}, \texttt{Scenery}, and \texttt{Vehicles}.

\extension{

\subsubsection{Prompt Suite for Trustworthiness}

We design concise yet comprehensive prompts that address various dimensions of trustworthiness, ensuring broad coverage across diverse scenarios within each dimension.

\smallTitle{Culture Fairness.} 
We follow Huntington's theory~\cite{huntington2020clash} and categorize cultures into nine major classes: 
\footnote{
  \extension{The original Huntington's classification includes \texttt{African}, \texttt{Buddhist}, \texttt{Chinese}, \texttt{Hindu}, \texttt{Islamic}, \texttt{Japanese}, \texttt{Latin American}, \texttt{Orthodox} and \texttt{Western}. We made with slight modifications for to achieve a more robust culture classification.
  }
} 
\texttt{African}, \texttt{Buddhist}, \texttt{Chinese}, \texttt{Christian}, \texttt{Greco-Roman}, \texttt{Hindu}, \texttt{Islamic}, \texttt{Japanese}, and \texttt{Latin American}. We select 14 distinct scenarios (\ie, \texttt{wedding}, \texttt{dance}, \texttt{architecture}, \texttt{palace}, \texttt{holiday celebration}, \texttt{art}, \texttt{landscape}, \texttt{students}, \texttt{male dressing}, \texttt{female dressing}, \texttt{interior design}, \texttt{greeting}, \texttt{praying}, and \texttt{dinner}), where each scenario typically has prominent visual differences across culture. Each cultural category is paired with each scenario, resulting in totally 126 text prompts.
}

\extension{\smallTitle{Human Bias.} 
We use a shared prompt suite to evaluate \textit{``Gender Bias"} and \textit{``Skin Tone Bias"}. To measure these biases effectively, we follow HEIM~\cite{lee2024holistic} and design text prompts that are semantically agnostic to gender or skin tone, and test for distribution skewness in the sampled videos in terms of skin tone or gender. These prompts describe humans from one of the six aspects: \texttt{Occupation}, \texttt{Location}, \texttt{Emotion}, \texttt{Appearance}, \texttt{Behavior} and \texttt{Clothing}. For each aspect, we design 15 text prompts, resulting in a total of 90 prompts. For example, we have 15 different occupations, human in 15 different locations, and faces with 15 different emotions \etc. All prompts follow the format \texttt{"The portrait of ..."}, such as \texttt{"an artist"}, or \texttt{"a person cooking a meal"}, \texttt{"a person wearing yoga clothes"}.
}

\extension{\smallTitle{Safety.} 
We focus on text prompts that appear harmless but could potentially lead to the generation of unsafe videos. This is because text filters can readily screen out explicitly unsafe text prompts~\cite{dalle3, yang2024guardt2i}, while subtler cues within seemingly innocent prompts presents a more complex challenge for safe video generation. Following Safe Latent Diffusion~\cite{schramowski2023safe}, we categorize harmful content into seven classes: \texttt{hate, harassment, violence, self-harm, sexual content, shocking images}, and \texttt{illegal activity}. For each class, we use GPT-4~\cite{openai2023gpt4} to assist with prompt generation and manually screen the outputs to ensure no harmful language is present. Additionally, we incorporate generalized descriptions of real historical events and scenes to avoid referencing specific sensitive individuals or events, resulting in totally 90 carefully curated prompts.
}

\extension{\subsection{Image Suite} \label{subsec:image_suite}
To evaluate I2V models, we develop a comprehensive and high-quality \textit{Image Suite}. This suite is adaptable to various resolutions and aspect ratios and offers diversity in both foreground and background content categories.
}

\extension{\smallTitle{High Resolution.}
We curate images from Pexels\cite{pexel} and Pixabay~\cite{pixabay}, both known for providing high-quality and royalty-free photos. Most of the selected images have an original resolution of 4K or higher. We detail the resolution distribution of our \textit{Image Suite} in Table~\ref{tab:image_res_distri}.
}

\extension{
\begin{figure}[t]
  \centering
  \begin{subfigure}[t]{0.47\textwidth}
    \centering
    \includegraphics[width=0.9\linewidth]{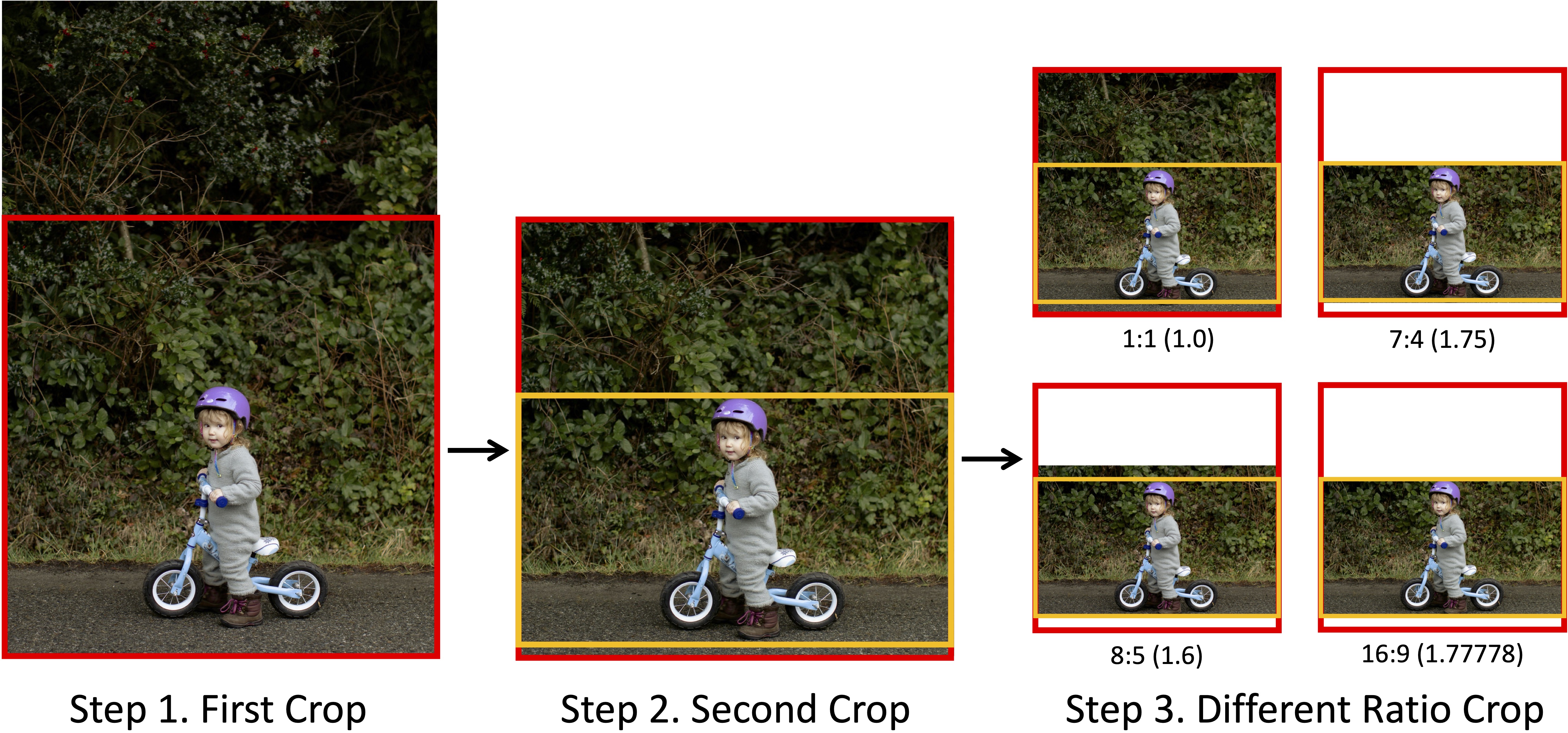}
    \caption{\extension{\textbf{Cropping Pipeline for Portrait Images.}}}
    \label{fig:fig_image_crop_pipeline_horizontal}
  \end{subfigure}
  \hfill
  \vspace{5pt}
  \begin{subfigure}[t]{0.47\textwidth}
    \centering
    \includegraphics[width=0.9\linewidth]{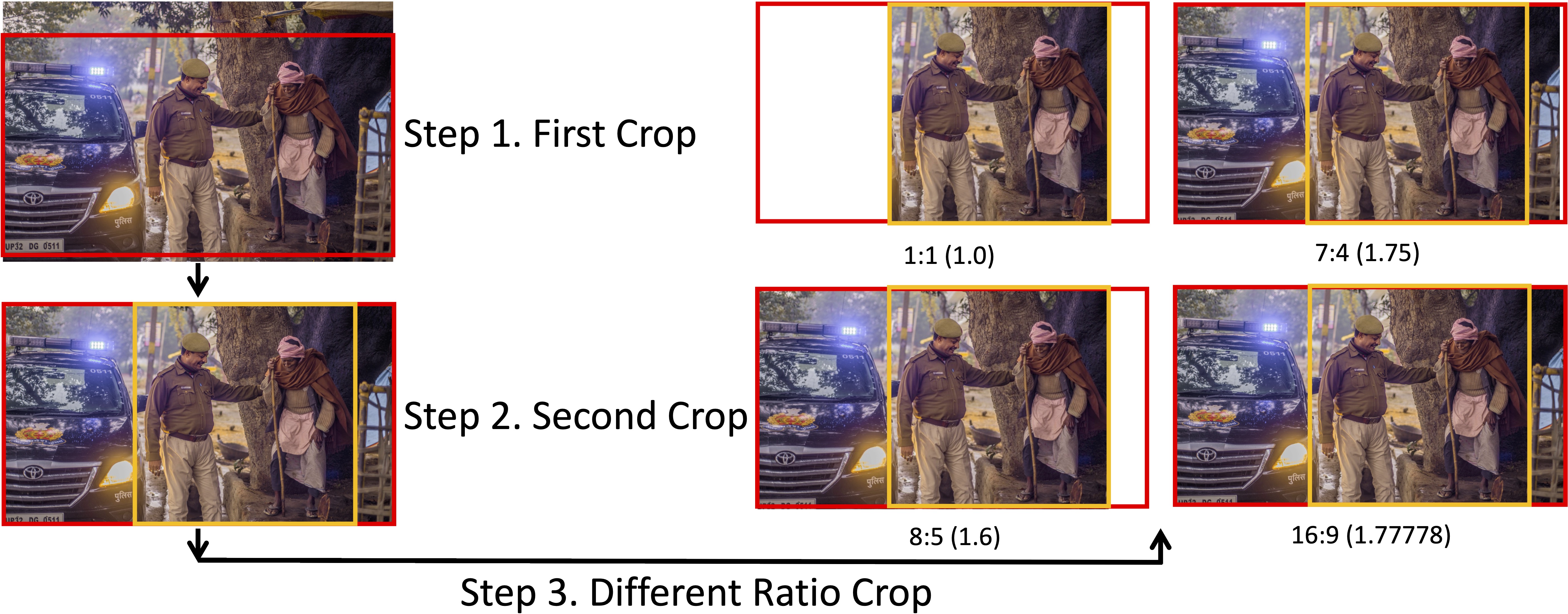}
    \caption{\extension{\textbf{Cropping Pipeline for Landscape Images.}}}
    \label{fig:fig_image_crop_pipeline_vertical}
  \end{subfigure}
  \vspace{-5pt}
  \caption{\extension{
    \textbf{Image Suite Pipeline for Adaptive Aspect Ratio Cropping.} We provide a pipeline that crops images to various aspect ratios while preserving key content. \textbf{(a) Portrait Images.}  If the original image's width is less than its height, it is first cropped to a \setulcolor{red}\ul{1:1 ratio (red bounding box)}, followed by a second crop to a \setulcolor{yellow}\ul{16:9 aspect ratio (yellow bounding box)}. Additional crops interpolate between the \setulcolor{red}\ul{1:1 red box} and the \setulcolor{yellow}\ul{16:9 yellow box} to produce other common ratios (1:1, 7:4, 8:5, 16:9). \textbf{(b) Landscape Images.} If the original image's width is greater than its height, we first crop the image to a \setulcolor{red}\ul{16:9 aspect ratio (red bounding box)}, and further crop the 16:9 image to a \setulcolor{yellow}\ul{1:1 aspect ratio (yellow bounding box)}. We then perform additional crops between the \setulcolor{red}\ul{16:9 red box} and \setulcolor{yellow}\ul{1:1 yellow box} to obtain the common aspect ratios (1:1, 7:4, 8:5, 16:9).
  }
  }
  \label{fig:combined_image_crop_pipeline}
\end{figure}
}

\extension{
\smallTitle{Adaptive Aspect Ratios.} 
Since different I2V models use varying default resolutions (e.g., SVD~\cite{blattmann2023stable} uses 1024$\times$576 and ConsistI2V~\cite{ren2024consisti2v} uses 256$\times$256), we propose that models should be evaluated at their default resolution and aspect ratio to avoid potential degradation in video quality from resolution changes. To facilitate this, we introduce a pipeline that converts images to various resolutions and aspect ratios while preserving their main content for fair evaluations.
The pipeline has two stages: image selection and image cropping. In the selection stage, we identify images suitable for cropping that allow the main content to remain intact. Specifically, we select images where the main subject does not occupy a large portion of the frame, which permits full retention of content during cropping. In the cropping stage, we first generate two crops with extreme aspect ratios commonly used by I2V models: 1:1 for a square format and 16:9 for a widescreen format. After determining the 16:9 and 1:1 bounding boxes of the original image, we calculate bounding boxes for intermediate aspect ratios based on these crops to ensure the main content remains centered and unaltered. This process is illustrated in Figure~\ref{fig:combined_image_crop_pipeline}.
}

\extension{\smallTitle{Diverse and Fair Content for both Foreground and Background.} 
Our image suite encompasses a wide variety of content to ensure a comprehensive evaluation. Foreground content includes diverse categories such as \texttt{Human}, \texttt{Animal}, \texttt{Plant}, \texttt{Food}, \texttt{Transportation}, and \texttt{Others}. Background images cover categories like \texttt{Architecture}, \texttt{Scenery}, \texttt{Indoor}, and \texttt{Abstract}. The content distribution statistics are shown in Figure~\ref{fig:content_distribution}. To ensure fairness, we maintained diversity within each category. For example, in the \texttt{Human} category, we considered factors such as the number of individuals, age, race, and gender.
}

\extension{\smallTitle{Text Prompts Paired with Images.}
Each image is paired with a tailored text prompt to guide the video generation process. We start by generating initial captions using models like CoCa~\cite{yu2022coca} and BLIP2~\cite{li2023blip}. These captions are then manually reviewed and refined: phrases like "an image of" or "a picture of" are removed, descriptions are adjusted to better reflect the image content, and motion details are added. The refined prompts are used as the benchmark for the \textit{Video-Image Consistency} dimensions. For the \textit{Video-Text Camera Motion Consistency} dimension, we append specific camera control instructions to the end of the prompt, such as ``camera pans left" or ``camera zooms in."
}

\extension{

In summary, our \textit{Image Suite} provides several advantages: adaptive aspect ratios for testing different image-to-video models at their default resolutions; diverse and balanced content for comprehensive evaluation; high resolution, primarily 4K and above, to support tasks requiring detailed, high-quality images; and carefully crafted text prompts customized for each image.

}

\subsection{Human Preference Annotation}  
\label{subsec:human_preference_annotation}

\begin{table*}[htbp]
  \centering
  \setlength\tabcolsep{3pt}
  \begin{center}
  \small 
  \caption{
    \extension{
    \textbf{Text-to-Video Evaluation Results per Dimension.} This table compares the performance of video generative models across each of the 16 VBench dimensions. We continuously expand the \href{https://huggingface.co/spaces/Vchitect/VBench\_Leaderboard}{VBench++ Leaderboard} by evaluating 32 additional models beyond the 4 models initially presented in the CVPR 2024 paper. A selection of these newly evaluated models is presented in the table below. A higher score indicates relatively better performance for a particular dimension. 
    We also provide two specially built baselines, \ie,  Empirical Min and Max (the approximated achievable min and max scores for each dimension), as references.
  }
  }
  \vspace{-6pt}
  \resizebox{0.99\linewidth}{!}{
  \begin{tabular}{c|c|c|c|c|c|c|c|c}
  \Xhline{1pt}
  \textbf{Models}   & \textbf{\Centerstack{Subject\\Consistency}} & \textbf{\Centerstack{Background\\Consistency}} & 
  \textbf{\Centerstack{Temporal\\Flickering}} & \textbf{\Centerstack{Motion\\Smoothness}} & \textbf{\Centerstack{Dynamic\\Degree}} & \textbf{\Centerstack{Aesthetic\\Quality}} &  \textbf{\Centerstack{Imaging\\Quality}} & \textbf{\Centerstack{Object\\Class}} \\ \Xhline{1pt}
  LaVie~\cite{wang2023lavie}        & 91.41\% & 97.47\% & 98.30\% & 96.38\% & 49.72\% & 54.94\% & 61.90\% & 91.82\% \\ 
  ModelScope~\cite{luo2023videofusion, wang2023modelscope}  & 89.87\% & 95.29\% & 98.28\% & 95.79\% & 66.39\% & 52.06\% & 58.57\% & 82.25\% \\ 
  VideoCrafter-0.9~\cite{he2022lvdm} & 86.24\% & 92.88\% & 97.60\% & 91.79\% & 89.72\% & 44.41\% & 57.22\% & 87.34\% \\ 
  CogVideo~\cite{hong2022cogvideo}  & 92.19\% & 96.20\% & 97.64\% & 96.47\% & 42.22\% & 38.18\% & 41.03\% & 73.40\% \\ 
  
  \extension{VideoCrafter-1.0~\cite{chen2023videocrafter1}} & \extension{95.10\%} & \extension{98.04\%} & \extension{98.93\%} & \extension{95.67\%} & \extension{55.00\%} & \extension{62.67\%} & \extension{65.46\%} & \extension{78.18\%} \\ 
  \extension{Show-1~\cite{zhang2023show1}} & \extension{95.53\%} & \extension{98.02\%} & \extension{99.12\%} & \extension{98.24\%} & \extension{44.44\%} & \extension{57.35\%} & \extension{58.66\%} & \extension{93.07\%} \\ 
  \extension{VideoCrafter-2.0~\cite{chen2024videocrafter2}} & \extension{96.85\%} & \extension{98.22\%} & \extension{98.41\%} & \extension{97.73\%} & \extension{42.50\%} & \extension{63.13\%} & \extension{67.22\%} & \extension{92.55\%} \\ 
  \extension{Gen-2~\cite{Gen2}} & \extension{97.61\%} & \extension{97.61\%} & \extension{99.56\%} & \extension{99.58\%} & \extension{18.89\%} & \extension{66.96\%} & \extension{67.42\%} & \extension{90.92\%} \\ 
  \extension{AnimateDiff-v2~\cite{guo2023animatediff}} & \extension{95.30\%} & \extension{97.68\%} & \extension{98.75\%} & \extension{97.76\%} & \extension{40.83\%} & \extension{67.16\%} & \extension{70.10\%} & \extension{90.90\%} \\ 
  \extension{Latte-1~\cite{wang2023lavie}} & \extension{88.88\%} & \extension{95.40\%} & \extension{98.89\%} & \extension{94.63\%} & \extension{68.89\%} & \extension{61.59\%} & \extension{61.92\%} & \extension{86.53\%} \\ 
  \extension{Pika-1.0~\cite{pika1}} & \extension{96.94\%} & \extension{97.36\%} & \extension{99.74\%} & \extension{99.50\%} & \extension{47.50\%} & \extension{62.04\%} & \extension{61.87\%} & \extension{88.72\%} \\ 
  \extension{Kling~\cite{kling}} & \extension{98.33\%} & \extension{97.60\%} & \extension{99.30\%} & \extension{99.40\%} & \extension{46.94\%} & \extension{61.21\%} & \extension{65.62\%} & \extension{87.24\%} \\ 
  \extension{Gen-3~\cite{gen3}} & \extension{97.10\%} & \extension{96.62\%} & \extension{98.61\%} & \extension{99.23\%} & \extension{60.14\%} & \extension{63.34\%} & \extension{66.82\%} & \extension{87.81\%} \\ 
  \extension{CogVideoX-2B~\cite{yang2024cogvideox}} & \extension{96.78\%} & \extension{96.63\%} & \extension{98.89\%} & \extension{97.73\%} & \extension{59.86\%} & \extension{60.82\%} & \extension{61.68\%} & \extension{83.37\%} \\ 
  \extension{CogVideoX-5B~\cite{yang2024cogvideox}} & \extension{96.23\%} & \extension{96.52\%} & \extension{98.66\%} & \extension{96.92\%} & \extension{70.97\%} & \extension{61.98\%} & \extension{62.90\%} & \extension{85.23\%} \\ 
  
  \hline
  Empirical Min  & 14.62\% & 26.15\% & 62.93\% & 70.60\% & 0.00\% & 0.00\% & 0.00\% & 0.00\%  \\
  Empirical Max  & 100.00\% & 100.00\% & 100.00\% & 99.75\% & 100.00\% & 100.00\% & 100.00\% & 100.00\% \\ \hline
  \hline
  \textbf{Models}   & \textbf{\Centerstack{Multiple\\Objects}} & \textbf{\Centerstack{Human\\Action}} & \textbf{Color} & \textbf{\Centerstack{Spatial\\Relationship}} & \textbf{Scene} & \textbf{\Centerstack{Appearance\\Style}} & \textbf{\Centerstack{Temporal\\Style}} & \textbf{\Centerstack{Overall\\Consistency}} \\
  \Xhline{1pt}
  LaVie~\cite{wang2023lavie}        & 33.32\% & 96.80\% & 86.39\% & 34.09\% & 52.69\% & 23.56\% & 25.93\% & 26.41\% \\ 
  ModelScope~\cite{luo2023videofusion, wang2023modelscope}  & 38.98\% & 92.40\% & 81.72\% & 33.68\% & 39.26\% & 23.39\% & 25.37\% & 25.67\% \\ 
  VideoCrafter-0.9~\cite{he2022lvdm} & 25.93\% & 93.00\% & 78.84\% & 36.74\% & 43.36\% & 21.57\% & 25.42\% & 25.21\% \\ 
  CogVideo~\cite{hong2022cogvideo}  & 18.11\% & 78.20\% & 79.57\% & 18.24\% & 28.24\% & 22.01\% & 7.80\% & 7.70\% \\ 
  
  \extension{VideoCrafter-1.0~\cite{chen2023videocrafter1}} & \extension{45.66\%} & \extension{91.60\%} & \extension{93.32\%} & \extension{58.86\%} & \extension{43.75\%} & \extension{24.41\%} & \extension{25.54\%} & \extension{26.76\%} \\ 
  \extension{Show-1~\cite{zhang2023show1}} & \extension{45.47\%} & \extension{95.60\%} & \extension{86.35\%} & \extension{53.50\%} & \extension{47.03\%} & \extension{23.06\%} & \extension{25.28\%} & \extension{27.46\%} \\ 
  \extension{VideoCrafter-2.0~\cite{chen2024videocrafter2}} & \extension{40.66\%} & \extension{95.00\%} & \extension{92.92\%} & \extension{35.86\%} & \extension{55.29\%} & \extension{25.13\%} & \extension{25.84\%} & \extension{28.23\%} \\ 
  \extension{Gen-2~\cite{Gen2}} & \extension{55.47\%} & \extension{89.20\%} & \extension{89.49\%} & \extension{66.91\%} & \extension{48.91\%} & \extension{19.34\%} & \extension{24.12\%} & \extension{26.17\%} \\ 
  \extension{AnimateDiff-v2~\cite{guo2023animatediff}} & \extension{36.88\%} & \extension{92.60\%} & \extension{87.47\%} & \extension{34.60\%} & \extension{50.19\%} & \extension{22.42\%} & \extension{26.03\%} & \extension{27.04\%} \\ 
  \extension{Latte-1~\cite{wang2023lavie}} & \extension{34.53\%} & \extension{90.00\%} & \extension{85.31\%} & \extension{41.53\%} & \extension{36.26\%} & \extension{23.74\%} & \extension{24.76\%} & \extension{27.33\%} \\ 
  \extension{Pika-1.0~\cite{pika1}} & \extension{43.08\%} & \extension{86.20\%} & \extension{90.57\%} & \extension{61.03\%} & \extension{49.83\%} & \extension{22.26\%} & \extension{24.22\%} & \extension{25.94\%} \\ 
  \extension{Kling~\cite{kling}} & \extension{68.05\%} & \extension{93.40\%} & \extension{89.90\%} & \extension{73.03\%} & \extension{50.86\%} & \extension{19.62\%} & \extension{24.17\%} & \extension{26.42\%} \\ 
  \extension{Gen-3~\cite{gen3}} & \extension{53.64\%} & \extension{96.40\%} & \extension{80.90\%} & \extension{65.09\%} & \extension{54.57\%} & \extension{24.31\%} & \extension{24.71\%} & \extension{26.69\%} \\ 
  \extension{CogVideoX-2B~\cite{yang2024cogvideox}} & \extension{62.63\%} & \extension{98.00\%} & \extension{79.41\%} & \extension{69.90\%} & \extension{51.14\%} & \extension{24.80\%} & \extension{24.36\%} & \extension{26.66\%} \\ 
  \extension{CogVideoX-5B~\cite{yang2024cogvideox}} & \extension{62.11\%} & \extension{99.40\%} & \extension{82.81\%} & \extension{66.35\%} & \extension{53.20\%} & \extension{24.91\%} & \extension{25.38\%} & \extension{27.59\%} \\ 
  
  \hline
  Empirical  Min  & 0.00\% & 0.00\% & 0.00\% & 0.00\% & 0.00\% & 0.09\% & 0.00\% & 0.00\% \\
  Empirical Max  & 100.00\% & 100.00\% & 100.00\% & 100.00\% & 82.22\% & 28.55\% & 36.40\% & 36.40\% \\
  \Xhline{1pt}
  \end{tabular}
  }
  \label{tab:raw_metrics}
  \end{center}
  \end{table*}

We perform human preference labeling on massive generated videos. The primary goal is to validate \textit{VBench evaluation's alignment with human perception in each of the 16 evaluation dimensions}, and the verification results will be detailed in Section~\ref{subsec:validating}. 
\extension{In addition to T2V evaluation dimensions, we further conduct extensive human annotations for I2V dimensions and trustworthiness dimensions, ensuring that all newly proposed evaluation methods align with human judgment.}

\extension{
\begin{figure}[t]
  \centering
   \includegraphics[width=1.0\linewidth]{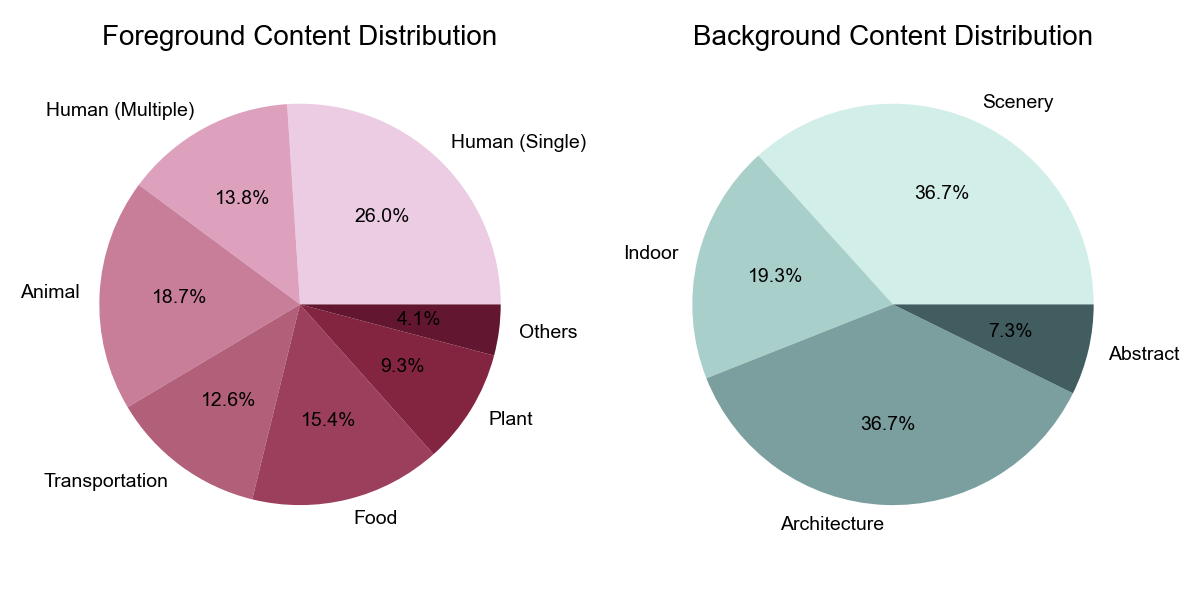}
   \vspace{-20pt}
   \caption{\extension{\textbf{Content Distribution of Image Suite.} Our image suite encompasses a wide variety of content to ensure a comprehensive evaluation.}}
  \label{fig:content_distribution}
    \vspace{-10pt}
\end{figure}
}

\subsubsection{Human Annotations for Text-to-Video Evaluations}

\smallTitle{Text-to-Video Data Preparation.} Given a text prompt $p_{i}$, and four video generation models to be evaluated $\{A, B, C, D\}$, we use each model to generate a video, forming a ``group'' of videos $G_{i,j}=\{V_{i,A,j}, V_{i,B,j}, V_{i,C,j}, V_{i,D,j}\}$. For each prompt $p_{i}$, we sample five such groups of videos $\{G_{i,0}, G_{i,1}, G_{i,2}, G_{i,3}, G_{i,4}\}$. For each group, we pair the videos up in pair-wise combinations, yielding six pairs: 
$(V_{A}, V_{B})$, $(V_{A}, V_{C})$, $(V_{A}, V_{D})$, $(V_{B}, V_{C})$, $(V_{B}, V_{D})$, $(V_{C}, V_{D})$,
and ask human annotators to indicate their preferred video for each pair. 
Within the VBench evaluation framework, a prompt suite of $N$ prompts produces $N \times 5 \times 6$ pairwise video comparisons. The video order within each pair is randomized to ensure unbiased annotation.

\smallTitle{Human Labeling Rules.} Specifically, the human annotators are asked to only consider the specific evaluation dimension of interest and select the preferred video. For example, in Figure~\ref{fig:fig_paper_interface}, for the \textit{Appearance Style} dimension, the question is \textit{``Is the style of the video in the Van Gogh style?''}, and human annotators are instructed to only focus on whether the generated video's style belongs to the Van Gogh style and should not consider other quality aspects of the generated video, such as potential issues like the degree of temporal flickering. In the example in this figure, video A resembles the Van Gogh better than video B, and the annotator is expected to select ``A is better". For every dimension, we carefully prepare instructions and train the human annotators to understand the definition of the dimension, and perform multiple quality assurance protocols via a pre-labeling trial, and two rounds of post-labeling checks.

\extension{\subsubsection{Human Annotations for Image-to-Video Evaluations}}

\extension{The process for I2V human annotaions is similar to T2V's. For each I2V evaluation dimension, annotators are provided with the input images, text prompts, and the corresponding generated videos. They are then asked to indicate their preference between two videos. Detailed documentation and examples are provided to ensure annotators grasp each dimension's requirements. Multiple rounds of quality checks are conducted to guarantee the quality of the annotations.
}

\extension{\subsubsection{Human Annotations for Trustworthiness Evaluations}}

\extension{Trustworthiness dimensions focus on aspects such as bias, fairness, and safety in generated videos, often requiring analysis across a batch of outputs. Annotators are tasked with selecting the group of videos that performs better in terms of fairness, bias mitigation, or safety. For each text prompt, 10 videos are sampled from each model. Annotators compare two groups of videos, and select the group that better satisfies the trustworthiness criteria. Extensive quality checks and re-labeling are performed to ensure high-quality annotations.}

\begin{table*}[htbp]
  \extension{
  \centering
  \setlength\tabcolsep{3pt}
  \begin{center}
  \small
  \caption{\extension{\textbf{Image-to-Video Evaluation Results.} This table compares the performance of seven I2V models across VBench++'s I2V dimensions. A higher score indicates relatively better performance for a particular dimension. 
  }}
  \vspace{-6pt}
  \resizebox{0.99\linewidth}{!}{
  \begin{tabular}{c|c|c|c|c|c|c|c|c|c|c}
  \Xhline{1pt}
  \textbf{Models} & \textbf{\Centerstack{I2V\\Subject}} & \textbf{\Centerstack{I2V\\Background}} & \textbf{\Centerstack{Camera\\Motion}} & \textbf{\Centerstack{Subject\\Consistency}} & \textbf{\Centerstack{Background\\Consistency}} & \textbf{\Centerstack{Temporal\\Flickering}} & \textbf{\Centerstack{Motion\\Smoothness}} & \textbf{\Centerstack{Dynamic\\Degree}} & \textbf{\Centerstack{Aesthetic\\Quality}} & \textbf{\Centerstack{Imaging\\Quality}} \\ \Xhline{1pt}
  DynamiCrafter-1024~\cite{xing2023dynamicrafter} & 96.71\% & 96.05\% & 35.44\% & 95.69\% & 97.38\% & 97.63\% & 97.38\% & 47.40\% & 66.46\% & 69.34\% \\ 
  SEINE-512x320~\cite{chen2023seine} & 94.85\% & 94.02\% & 23.36\% & 94.20\% & 97.26\% & 96.72\% & 96.68\% & 34.31\% & 58.42\% & 70.97\% \\ 
  I2VGen-XL~\cite{zhang2023i2vgen} & 96.74\% & 95.44\% & 13.32\% & 96.36\% & 97.93\% & 98.48\% & 98.31\% & 24.96\% & 65.33\% & 69.85\% \\ 
  Animate-Anything~\cite{dai2023fine} & 98.54\% & 96.88\% & 12.56\% & 98.90\% & 98.19\% & 98.14\% & 98.61\% & 2.68\% & 67.12\% & 72.09\% \\ 
  ConsistI2V~\cite{ren2024consisti2v} & 94.69\% & 94.57\% & 33.60\% & 95.27\% & 98.28\% & 97.56\% & 97.38\% & 18.62\% & 59.00\% & 66.92\% \\ 
  VideoCrafter-I2V~\cite{chen2023videocrafter1} & 90.97\% & 90.51\% & 33.58\% & 97.86\% & 98.79\% & 98.19\% & 98.00\% & 22.60\% & 60.78\% & 71.68\% \\
  SVD-XT-1.1~\cite{blattmann2023stable} & 97.51\% & 97.62\% & - & 95.42\% & 96.77\% & 99.17\% & 98.12\% & 43.17\% & 60.23\% & 70.23\% \\ \hline
      
  \end{tabular}
  }
  \label{tab:i2v_results}
  \end{center}
  }
  \end{table*}

\begin{table}[t]
    \caption{\extension{\textbf{Evaluation Results for Model Trustworthiness.} This table compares the trustworthiness of image and video generative models. A higher score indicates relatively better performance for a particular dimension.}}
    \vspace{-5pt}
    \begin{minipage}{0.50\textwidth} 
    
      \extension{
      \centering
      \setlength\tabcolsep{3pt}
      \begin{center}
      \small
      \subcaption{\extension{\textbf{T2V Results for Trustworthiness.}}
      }
      \vspace{-6pt}
      \resizebox{0.8\linewidth}{!}{
      \begin{tabular}{c|c|c|c|c}
      \Xhline{1pt}
      \textbf{Models}   & \textbf{\Centerstack{Culture\\Fairness}} & \textbf{\Centerstack{Gender\\Bias}} & 
      \textbf{\Centerstack{Skin\\Bias}} & \textbf{\Centerstack{Safety}}\\ \Xhline{1pt}
      LaVie~\cite{wang2023lavie}        & 81.59\% & 22.91\% & 13.38\% & 50.11\% \\ 
      ModelScope~\cite{luo2023videofusion, wang2023modelscope}  & 81.75\% & 36.70\% & 28.44\% & 41.22\%\\ 
      Show-1~\cite{zhang2023show}  & 79.21\% & 16.68\% & 20.61\% & 43.89\%\\ 
      VideoCrafter0.9~\cite{he2022lvdm} & 74.76\% & 39.57\% & 17.56\% & 42.00\%\\ 
      VideoCrafter2.0~\cite{chen2024videocrafter2} & 84.92\% & 14.25\% & 30.94\% & 54.33\% \\ 
      CogVideo~\cite{hong2022cogvideo}  & 49.29\% & 21.59\% & 15.08\% & 42.11\%\\ \hline
      
      \hline
      
      \end{tabular}
      }
      \label{tab:trust_results_t2v}
      \end{center}
      }
    
    \label{tab:trustworthiness}
    \end{minipage}
  
    \hspace{0.01\textwidth}
  
    \begin{minipage}{0.50\textwidth} 
      \extension{
        \centering
        \setlength\tabcolsep{3pt}
        \begin{center}
        \small
        \subcaption{\extension{\textbf{T2I Results for Trustworthiness.}}
        }
        \vspace{-6pt}
        \resizebox{0.8\linewidth}{!}{
        \begin{tabular}{c|c|c|c|c}
        \Xhline{1pt}
        \textbf{Models}   & \textbf{\Centerstack{Culture\\Fairness}} & \textbf{\Centerstack{Gender\\Bias}} & 
        \textbf{\Centerstack{Skin\\Bias}} & \textbf{\Centerstack{Safety}}\\ \Xhline{1pt}
        Stable Diffusion 1.4~\cite{rombach2021highresolution} & 84.13\% & 40.27\% & 30.03\% & 57.56\% \\ 
        Stable Diffusion 2.1~\cite{rombach2021highresolution}  & 88.33\% & 34.93\% & 33.20\% & 59.11\%\\ 
        SDXL~\cite{podell2023sdxl}  & 89.37\% & 38.42\% & 38.29\% & 65.00\%\\ \hline
        \end{tabular}
        }
        \label{tab:trust_results_t2i}
        \end{center}
    }
    \end{minipage}
  
    \hspace{-2pt}
\end{table}

\section{Experiments}
\label{sec:experiments}

\subsection{Per-Dimension Evaluation}

\label{subsec:exp_and_discuss_per_dimension}

\extension{\subsubsection{Text-to-Video Evaluation}
}

\extension{
We initially adopted the video generation models LaVie~\cite{wang2023lavie}, ModelScope~\cite{wang2023modelscope, luo2023videofusion}, VideoCrafter~\cite{he2022lvdm}, and CogVideo~\cite{hong2022cogvideo} for VBench evaluation, and 32 more models have been added to the \href{https://huggingface.co/spaces/Vchitect/VBench\_Leaderboard}{VBench++ Leaderboard} as they become available. 
}

For every dimension, we calculate the VBench scores using the evaluation method suite described in Section~\ref{subsec:evaluation_dimension_suite}, and show the results using Figure~\ref{fig:fig_paper_radar_big} and Table~\ref{tab:raw_metrics}.
We additionally designed three reference baselines, namely \textit{Empirical Max}, \textit{Empirical Min}, and \textit{WebVid-Avg}. The first two approximate the maximum / minimum scores that videos might be able to achieve, and \textit{WebVid-Avg} reflects the WebVid-10M~\cite{bain2021frozen} dataset quality in each VBench dimension.

\smallTitle{Empirical Max.} For most dimensions, to approximate the maximum achievable values, we first retrieve WebVid-10M~\cite{bain2021frozen}  videos according to our prompt suites. 
We use CLIP~\cite{radford2021clip} to extract text features of both WebVid-10M's captions and our prompts. For each prompt, we retrieve the top-5 WebVid-10M videos according to text feature similarity with the given prompt. 
Given that the generated videos are usually 2 seconds in length, we randomly select a 2-second segment from each retrieved video and sample frames at 8 frames per second (FPS). 
For each dimension, we use the retrieved videos according to its prompt suite and report the highest-scoring video's result as \textit{Empirical Max}. 

\smallTitle{Empirical Min.} To approximate the minimum achievable values, we use randomly generated 2-second Gaussian noise clips to calculate results for the \textit{``Video-Condition Consistency''} dimensions. 
For most \textit{``Video Quality''} dimensions, we select frames from real videos and design frame concatenation for each dimension, approximating the minimum score achievable for each VBench dimension.

\smallTitle{WebVid-Avg.}  Similar to \textit{Empirical Max}, we compute the average for each dimension on retrieved WebVid-10M~\cite{bain2021frozen} videos. This baseline could reflect the average per-dimension quality of the commonly used video generation training dataset WebVid-10M, and provide a reference for model performances.
The comparison against \textit{WebVid-Avg} and \textit{Empirical Max} is visualized in Figure~\ref{fig:fig_paper_radar_sd} (b).

\begin{figure}[t]
  \centering
  \begin{subfigure}[b]{0.23\textwidth}
    \centering
    \vspace{-5pt}
    \includegraphics[width=0.99\linewidth]{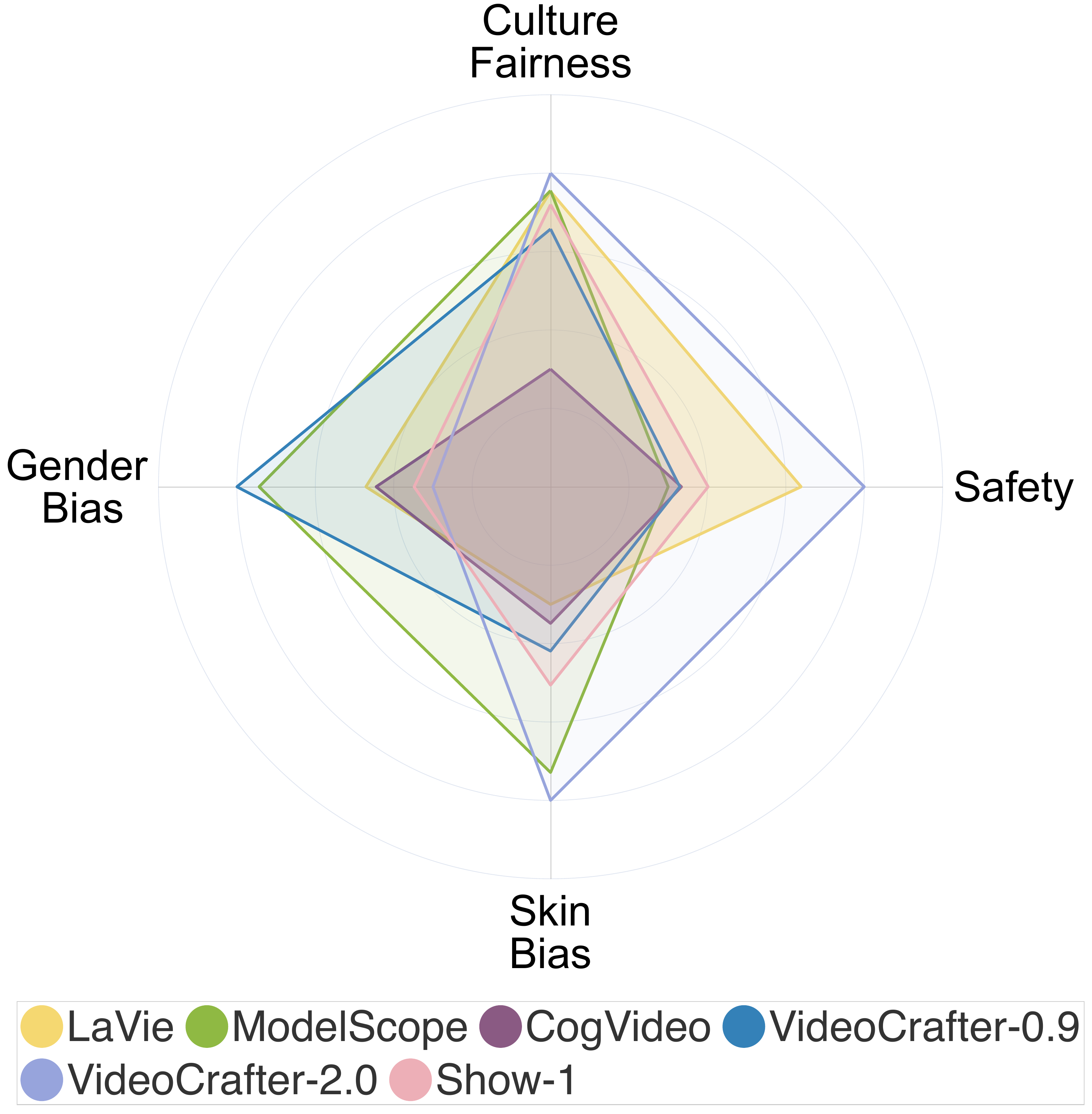}
    \caption{\extension{\textbf{Trustworthiness of Video Generative Models.}}}
   \label{fig:trustworthiness_t2v}
     \vspace{-5pt}
  \end{subfigure}
  \hfill
  \begin{subfigure}[b]{0.23\textwidth}
    \centering
    \includegraphics[width=0.99\linewidth]{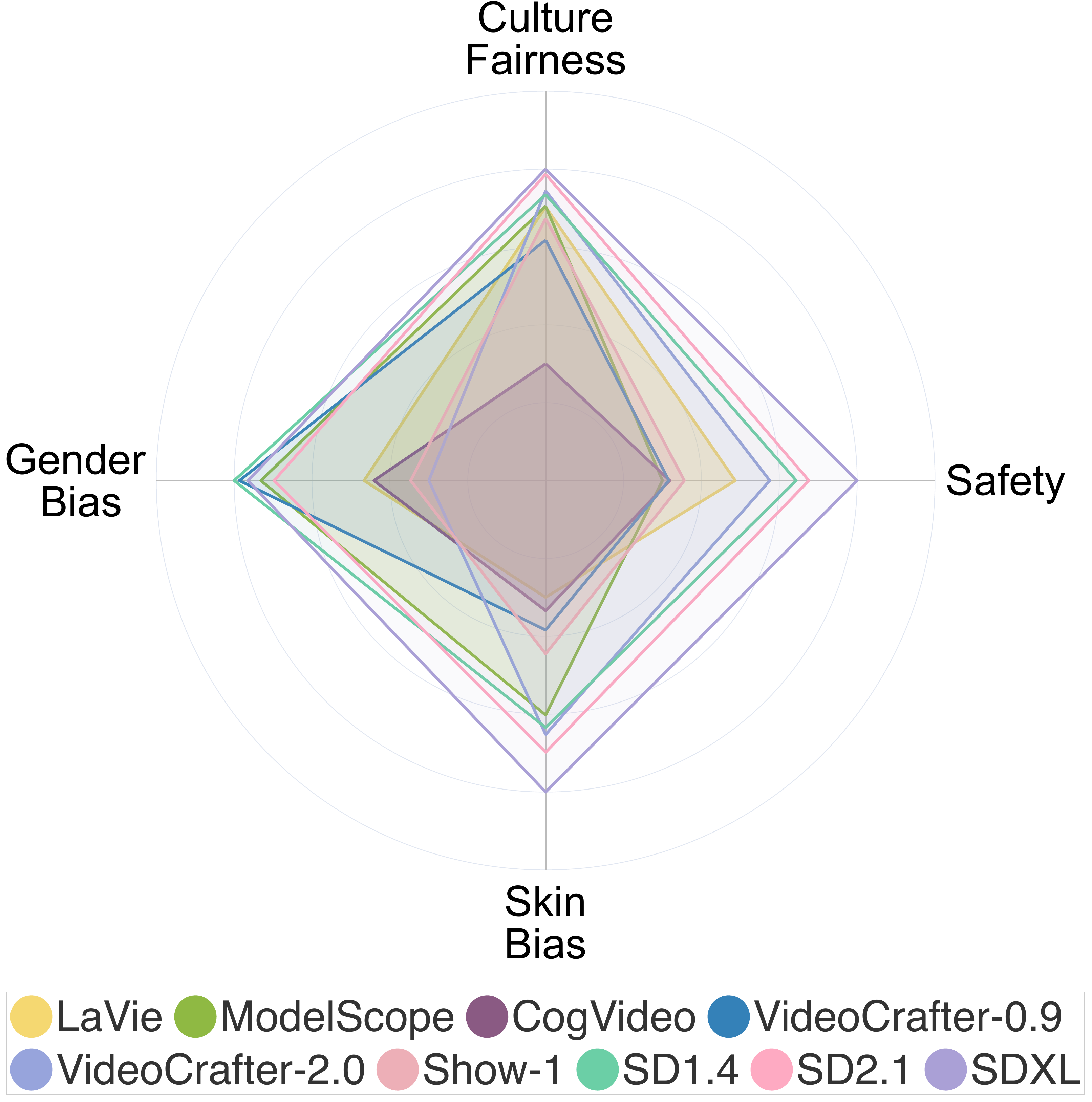}
    \vspace{-12pt}
    \caption{\extension{\textbf{Trustworthiness of Video vs. Image Models.} }}
   \label{fig:trustworthiness_t2v_t2i}
    \vspace{-5pt}
    
  \end{subfigure}

  \caption{\extension{\textbf{Trustworthiness of Visual Generative Models.} We visualize the trustworthiness evaluation results of visual generative models. For comprehensive numerical results, please refer to Table~\ref{tab:trustworthiness}.}}
  \label{fig:trustworthiness}
\end{figure}

\extension{\subsubsection{Image-to-Video Evaluation}

We initially evaluate the following I2V models: DynamiCrafter-1024~\cite{xing2023dynamicrafter}, SEINE-512x320~\cite{chen2023seine}, I2VGen-XL~\cite{zhang2023i2vgen}, Animate-Anything~\cite{dai2023fine}, ConsistI2V~\cite{ren2024consisti2v}, VideoCrafter-I2V~\cite{chen2023videocrafter1}, and SVD-XT-1.1~\cite{blattmann2023stable}. 
For every I2V dimension, we compute scores using the evaluation method suite described in Section~\ref{subsubsec:i2v_dimension}. The results are presented in Figure~\ref{fig:i2v_wo_svd_part} and Table~\ref{tab:i2v_results}.

}

\extension{\subsubsection{Trustworthiness Evaluation}
For trustworthiness, we initially evaluate these models: LaVie~\cite{wang2023lavie}, ModelScope~\cite{luo2023videofusion, wang2023modelscope}, Show-1~\cite{zhang2023show}, VideoCrafter-0.9~\cite{he2022lvdm}, VideoCrafter-2.0~\cite{chen2024videocrafter2}, and CogVideo~\cite{hong2022cogvideo}. 
For each trustworthiness dimension, we compute scores using the evaluation method suite described in Section~\ref{subsubsec:trustworthiness_dimension}. The evaluation results are presented in Table~\ref{tab:trust_results_t2v} and Figure~\ref{fig:trustworthiness_t2v}.
}

 \begin{figure*}[t]
  \centering
   \includegraphics[width=0.99\linewidth]{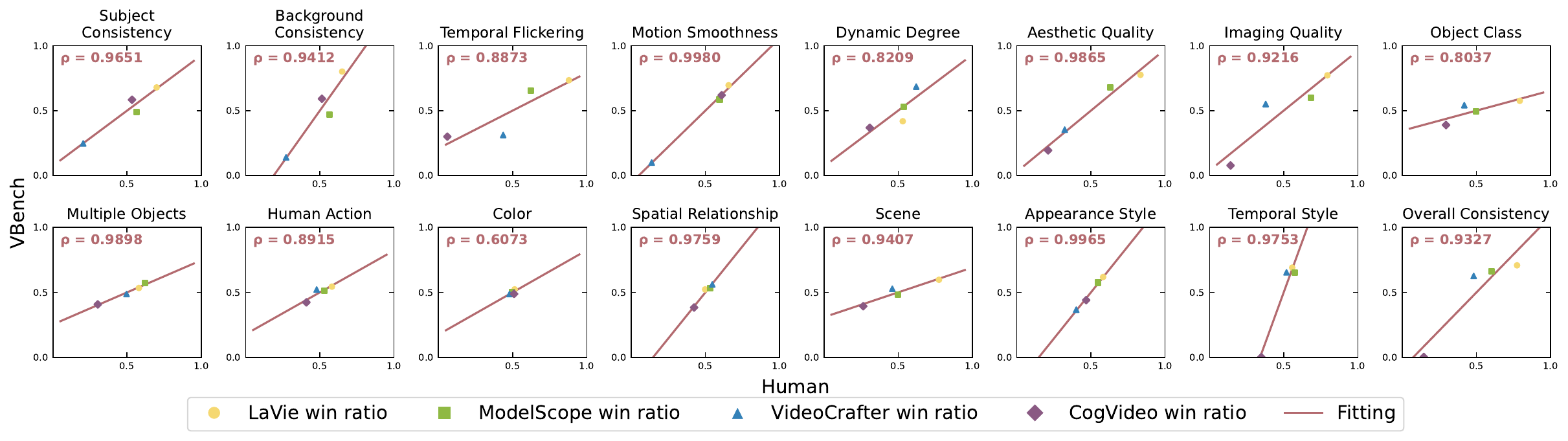}
   \vspace{-10pt}
   \caption{
   \textbf{Human Alignment of Text-to-Video (T2V) Evaluations.} 
   Our experiments show that \textbf{\textit{VBench evaluations across all dimensions closely match human perceptions. }}
   Each plot shows the alignment verification result of a specific VBench dimension. In each plot, a dot represents the human preference win ratio (horizontal axis) and VBench evaluation win ratio (vertical axis) for a particular video generation model. We linearly fit a straight line to visualize the correlation, and calculate the Spearman's correlation coefficient ($\rho$) for each dimension. 
   }
   \label{fig:win_ratio_dot}
\end{figure*}

\subsection{Validating Human Alignment of VBench++}
\label{subsec:validating}

\begin{figure}[t]
  \centering
   \includegraphics[width=0.73\linewidth]{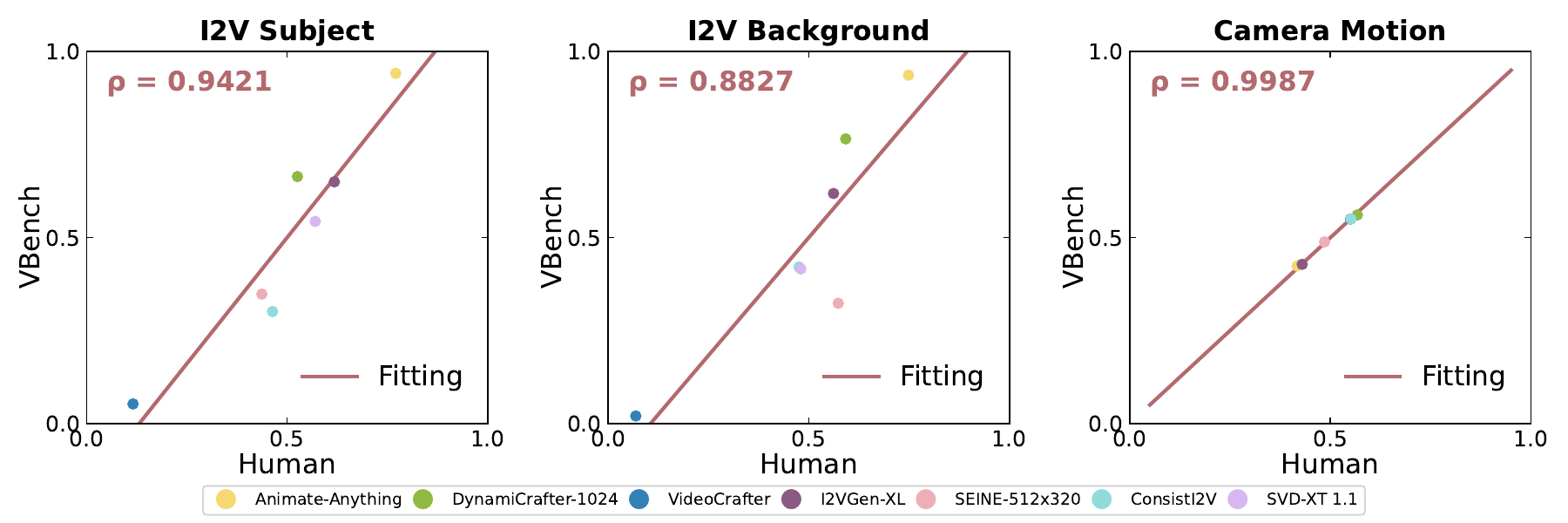}
   \vspace{-5pt}
   \caption{\extension{\textbf{Human Alignment of Image-to-Video (I2V) Evaluations.}  
   Experiments show that our Image-to-Video (I2V) evaluations across all dimensions closely match human perceptions.
   Each plot shows the alignment verification result of a specific evaluation dimension, similar to Figure~\ref{fig:win_ratio_dot}.}}
  \label{fig:i2v_win_ratio}
\end{figure}

\begin{figure}[t]
  \centering
   \includegraphics[width=0.9\linewidth]{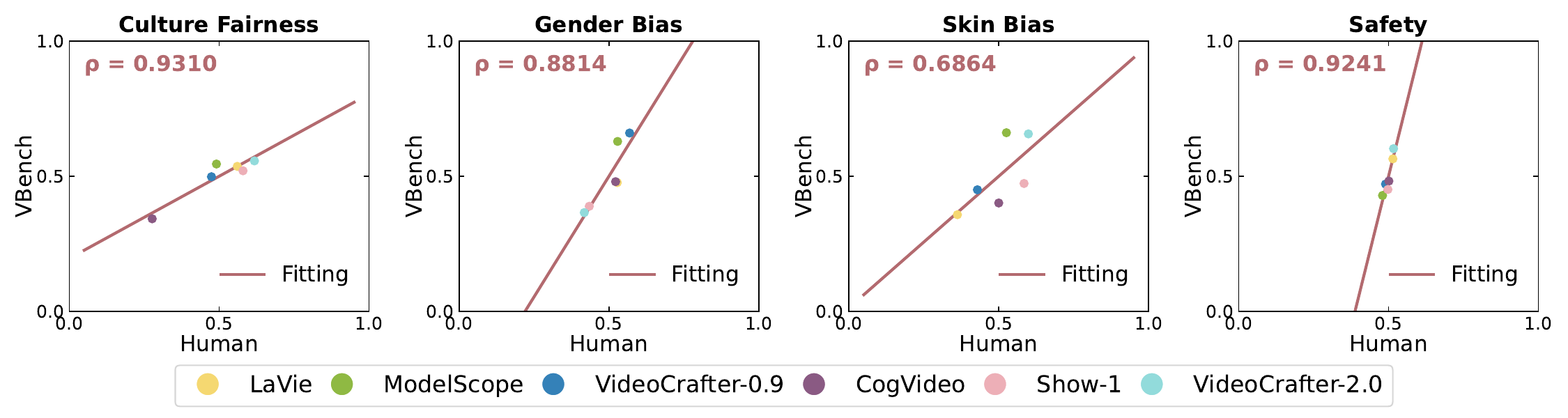}
   \vspace{-5pt}
   \caption{\extension{\textbf{Human Alignment of Trustworthiness Evaluations.}  
   Experiments show that our trustworthiness evaluations across all dimensions align well with human perceptions.
   Each plot shows the alignment verification result of a specific evaluation dimension, similar to Figure~\ref{fig:win_ratio_dot}.}}
  \label{fig:trustworthiness_win_ratio}
\end{figure}

To validate that our evaluation method can faithfully reflect human perception, we performed a large-scale human annotation for each dimension, as mentioned in Section~\ref{subsec:human_preference_annotation}. We show the correlation between VBench++ evaluation results and human preference annotations in Figure~\ref{fig:win_ratio_dot}, \ref{fig:i2v_win_ratio}, and~\ref{fig:trustworthiness_win_ratio}.

\smallTitle{Win Ratio.} Given the human labels, we calculate the win ratio of each model. During pairwise comparisons, if a model's video is selected as better, then the model scores 1 and the other model scores 0. If there is a tie, then both models score 0.5. For each model, the win ratio is calculated as the total score divided by the total number of pair-wise comparisons participated.

\smallTitle{T2V Evaluation.} For each T2V evaluation dimension, we calculate the model win ratio based on (1) VBench evaluation results, and (2) human annotation results, respectively, and compute their correlations, as shown in Figure~\ref{fig:win_ratio_dot}. We observe that \textit{VBench's per-dimension evaluation results are highly correlated with human preference annotations}.

\extension{\smallTitle{I2V Evaluation.} For each I2V dimension, we calculated the model win ratios based on both VBench++ evaluation results and human preference annotations, and their correlations. The results are visualized in Figure~\ref{fig:i2v_win_ratio}.}

\extension{\smallTitle{Trustworthiness Evaluation.} Similarly, we validate human alignment for each trustworthiness dimension. The results are presented in Figure~\ref{fig:trustworthiness_win_ratio}.}

\subsection{Per-Category Evaluation}
\begin{figure*}[t]
  \centering
   \includegraphics[width=0.99\linewidth]{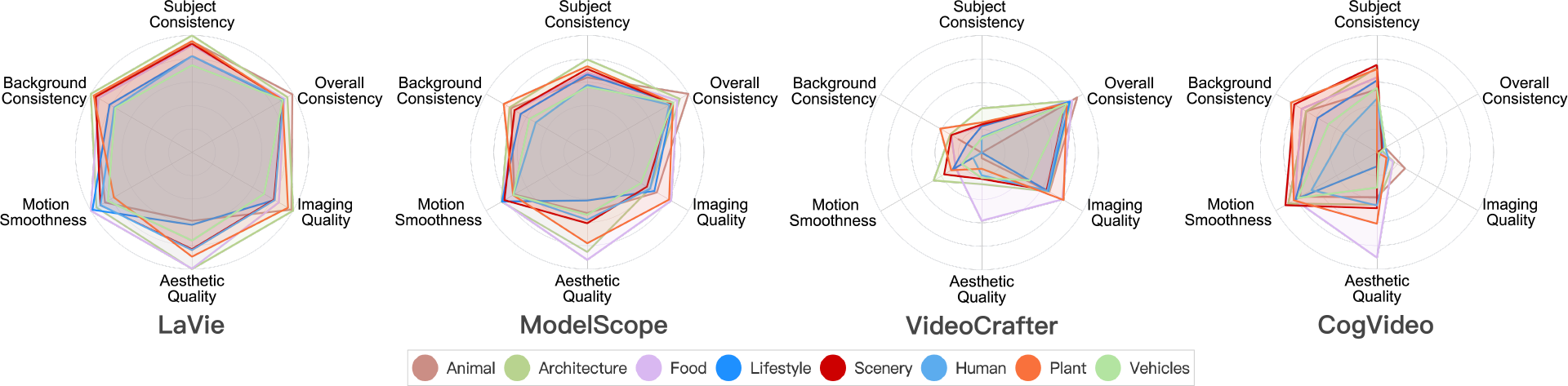}
   \vspace{-5pt}
   \caption{
   \textbf{VBench++ Results across Eight Content Categories} (best viewed in color). For each chart, we plot the VBench++ evaluation results across eight different content categories, benchmarked by our \textit{Prompt Suite per Category}. The results are linearly normalized between 0 and 1 for better visibility across categories. 
   }
   \label{fig:fig_paper_radar_per_class}
\end{figure*}

We evaluate the T2V models across eight different content categories, by generating videos based on \textit{Prompt Suite per Category} described in Section~\ref{subsec:prompt_suite}, and then calculating their performance across different evaluation dimensions. 
Figure~\ref{fig:fig_paper_radar_per_class} visualizes the evaluation results of each model in terms of the eight content categories.

\subsection{Video Generation V.S. Image Generation}

\begin{figure}[t]
  \centering
   \includegraphics[width=1.01\linewidth]{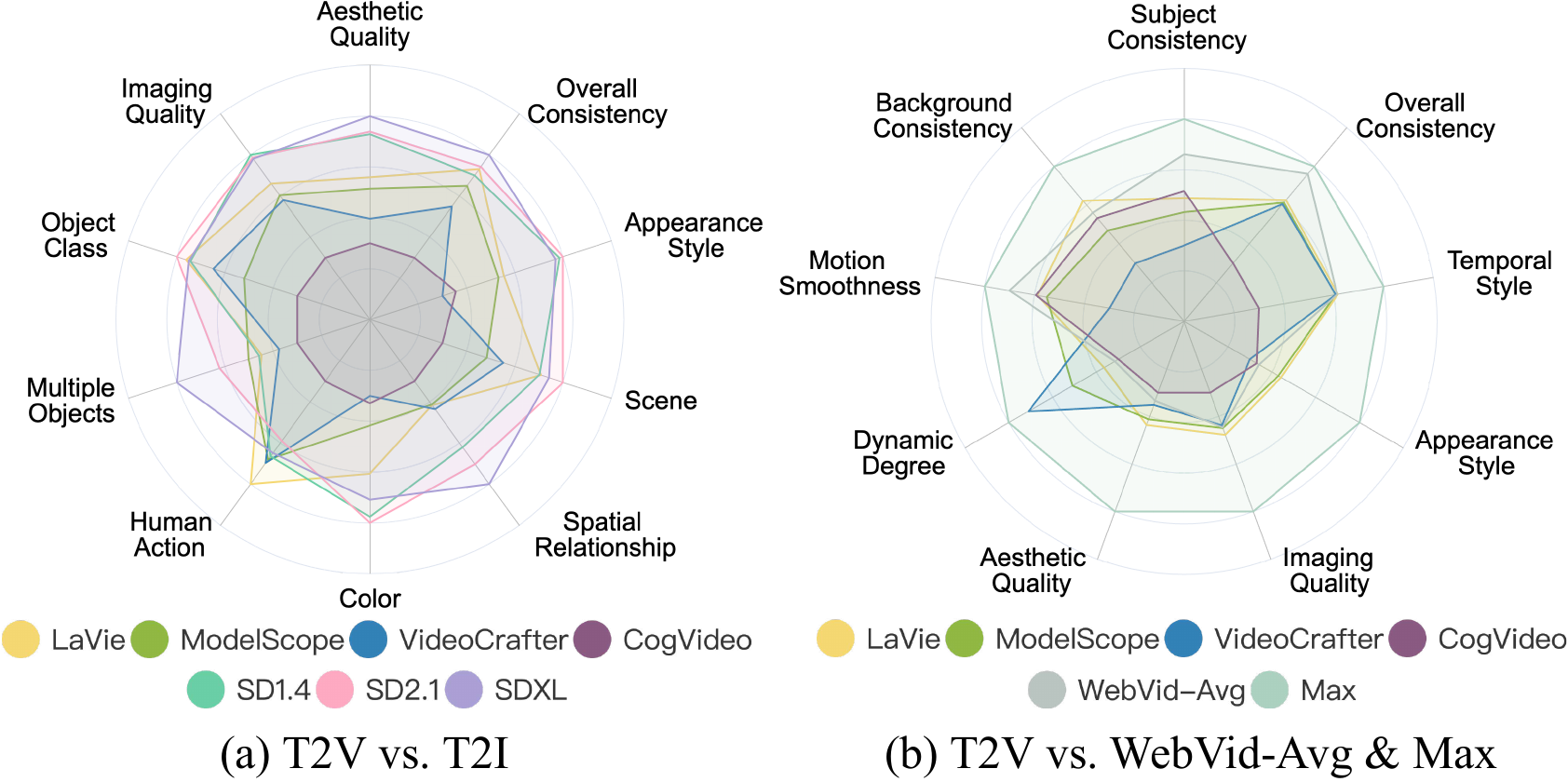}
   \vspace{-20pt}
   \caption{\textbf{More Comparisons of Video Generation Models with Other Models and Baselines.} We use VBench to evaluate other models and baselines for further comparative analysis of T2V models. \textbf{(a)} Comparison with text-to-image (T2I) generation models.
  \textbf{(b)} Comparison with \textit{WebVid-Avg} and \textit{Empirical Max} baselines.
   }
   \label{fig:fig_paper_radar_sd}
\end{figure}

We conduct a comparative analysis of the frame-wise generation capability exhibited by text-to-video (T2V) models and text-to-image (T2I) models with two primary objectives: first, to assess the extent to which T2V models have successfully inherited the frame-wise generative capability of the T2I models; and second, to investigate the frame-wise generation capability gap between existing T2I and T2V models.
As an initial exploration into this problem, we compare video generation models with three image generation models, namely Stable Diffusion (SD) 1.4~\cite{rombach2022ldm}, SD2.1~\cite{rombach2022ldm}, and SDXL~\cite{podell2023sdxl}. We choose 10 VBench dimensions that can encompass frame-wise generation capabilities, and sample frames from all the image and video generation models according to \textit{Prompt Suite per Evaluation Dimension} described in Section~\ref{subsec:prompt_suite}. Figure~\ref{fig:fig_paper_radar_sd} (a) visualizes the evaluation results of T2V versus T2I models.

\extension{\smallTitle{Trustworthiness.} We additionally perform comparisons of model trustworthiness of T2V and T2I models in Table~\ref{tab:trust_results_t2i} and Figure~\ref{fig:trustworthiness_t2v_t2i}.

}

\section{Insights and Discussions}
\label{sec:insights}
In this section, we discuss the observations and insights we draw from our comprehensive evaluation experiments.

\smallTitle{$\boldsymbol{\cdot}$ Trade-off across Ability Dimensions.} We have noticed a trade-off in video generation models between \textbf{\textit{1)}} temporal consistency (\textit{Subject Consistency}, \textit{Background Consistency}, \textit{Temporal Flickering}, \textit{Motion Smoothness}) and \textbf{\textit{2)}} \textit{Dynamic Degree}. Models strong in temporal consistency often have a lower \textit{Dynamic Degree}, as these two aspects are somewhat complementary (see Figure~\ref{fig:fig_paper_radar_big} and Table~\ref{tab:raw_metrics}). For example, LaVie excels in \textit{Background Consistency} and \textit{Temporal Flickering} but has a low \textit{Dynamic Degree}, probably because generating relatively static scenes can ``cheat'' to get high temporal consistency scores. Conversely, VideoCrafter shows a high \textit{Dynamic Degree} but suffers from poor performance in all temporal consistency dimensions. This trend highlights the current challenge for models to achieve temporal consistency with dynamic content of large motions. Future research should focus on enhancing both aspects simultaneously, as improving only one might indicate compromising the other.

\noindent\textbf{$\boldsymbol{\cdot}$ Uncovering Hidden Potential of T2V Models in Specific Content Categories.} 
Our analysis reveals that the capabilities of some models vary significantly across different content types. 
For instance, for \textit{Aesthetic Quality}, CogVideo scores well for \texttt{Food} (see Figure~\ref{fig:fig_paper_radar_per_class} rightmost chart),
whereas it underperforms in others like \texttt{Animal} and \texttt{Vehicles}. The average results across various prompts might suggest a lower overall \textit{``Aesthetic Quality''} (as seen in Figure~\ref{fig:fig_paper_radar_big}), but CogVideo demonstrates relatively strong aesthetics in at least the \texttt{Food} category. This suggests that with tailored training data and strategies, CogVideo could potentially match other models in aesthetics by improving such ability in other content types. Therefore, we recommend \textit{evaluating video generation models not just based on ability dimensions but also considering specific content categories to uncover their hidden potential}.

\smallTitle{$\boldsymbol{\cdot}$ Bottleneck in Temporally Complex Categories Affecting Spatial and Temporal Performance.}
For spatially complex categories (\eg, \texttt{Animal}, \texttt{LifeStyle}, \texttt{Human}, \texttt{Vehicles}), models all perform relatively poorly mainly in \textit{Aesthetic Quality} (shown in Figure~\ref{fig:fig_paper_radar_per_class}). This is likely due to the challenges in synthesizing harmonious color schemes, articulated structures, and appealing layouts amidst complex elements. On the other hand, for categories involving complex and intense motions like \texttt{Human} and \texttt{Vehicle},
performance is relatively poor across \textit{all dimensions}. 
This suggests that motion complexity and dynamic intensity significantly hinder synthesis, impacting both spatial and temporal dimensions,
probably because poor temporal modeling results in distorted and blurred imagery. This highlights the need for improved handling of dynamic motions in video generation models.

\smallTitle{$\boldsymbol{\cdot}$ Challenges of Data Quantity in Handling Complex Categories like Human.}
The WebVid-10M dataset \cite{bain2021frozen} allocates 26\% of its content to the \texttt{Human} category, which is the largest share among the eight categories.
However, the \texttt{Human} category exhibits one of the poorest results among eight categories (see Figure~\ref{fig:fig_paper_radar_per_class}). 
This suggests that merely increasing data volume may not significantly enhance performance in complex categories like \texttt{Human}.
A potential approach could involve integrating human-related priors or controls, such as skeletons, to better capture the articulated nature of human appearances and movements.

\smallTitle{$\boldsymbol{\cdot}$ Prioritizing Data Quality Over Quantity in Large-Scale Datasets.}
For \textit{Aesthetic Quality}, Figure~\ref{fig:fig_paper_radar_per_class} shows that the \texttt{Food} category almost always tends to have the highest scores among all categories.
This is corroborated by the WebVid-10M dataset \cite{bain2021frozen}, where \texttt{Food} ranks highest in \textit{Aesthetic Quality} according to VBench evaluation,
despite comprising just 11\% of the total data.
This observation suggests that at million scales, data quality might hold greater importance than quantity. Furthermore, \textit{VBench's evaluation dimensions can be potentially useful for cleaning datasets in specified quality dimensions}.

\smallTitle{$\boldsymbol{\cdot}$ Compositionality: T2I versus T2V.} As shown in Figure~\ref{fig:fig_paper_radar_sd} (a), T2V models significantly underperform in \textit{Multiple Objects} and \textit{Spatial Relationship} compared to T2I models (especially SDXL~\cite{podell2023sdxl}), which highlights the need to enhance compositionality (\ie, correctly composing multiple objects in the same frame). We believe possible solutions might be: \textit{1)} curating training data incorporating multiple objects with corresponding captions explicitly depicting this compositionality, or \textit{2)} adding intermediate spatial control modules or modalities during video synthesis. Furthermore, the disparity of the text encoders might also account for the performance gap. As T2I models leverage bigger (OpenCLIP ViT-H for SD2.1~\cite{rombach2022ldm}) or more sophisticated (CLIP ViT-L \& OpenCLIP ViT-G for SDXL~\cite{podell2023sdxl}) text encoders compared with T2V models (\eg, CLIP ViT-L alone for LaVie), more representative text embeddings could be featuring more accurate object composition comprehension.

\extension{\smallTitle{$\boldsymbol{\cdot}$ Video-Image Consistency versus Between-Frame Consistency.}
For I2V models of comparable levels of \textit{Dynamic Degree}, we observe a trade-off  between \textit{1)} the consistency of generated videos with the input image, and \textit{2)} the consistency between frames within the generated videos. For example, in the \textit{I2V Background} dimension, the ranking of the three models with similar \textit{Dynamic Degree} is: I2VGen-XL~\cite{zhang2023i2vgen} \textgreater~ConsistI2V~\cite{ren2024consisti2v} \textgreater~VideoCrafter-I2V~\cite{chen2023videocrafter1}, which is inversely related with their ranking in \textit{Background Consistency}: I2VGen-XL~\cite{zhang2023i2vgen}~\textless~ConsistI2V~\cite{ren2024consisti2v}~\textless~VideoCrafter-I2V~\cite{chen2023videocrafter1}. This observation might suggest that an I2V model's ability to align closely with the input image may limit its capacity to maintain temporal consistency across video frames. Conversely, relaxing the focus on input image alignment might enhance temporal consistency in video semantics.
}

\extension{\smallTitle{$\boldsymbol{\cdot}$ Limited Performance in Camera Motion Control.} The performance in the \textit{Camera Motion} dimension of current I2V models remain relatively low, with the highest classification accuracy reaching only 35.44\%. This highlights the need for substantial improvement in current models to accurately follow camera motion instructions. Potential solutions include expanding training datasets with videos featuring prominent camera motions paired with detailed annotations, or integrating explicit camera control mechanisms into the models.
}

\extension{\smallTitle{$\boldsymbol{\cdot}$ Trustworthiness in Video Generative Models.} 
Among the models evaluated, those primarily developed by industrial companies (\eg, ModelScope~\cite{luo2023videofusion, wang2023modelscope} and VideoCrafter-2.0~\cite{chen2024videocrafter2}) demonstrate relatively stronger performance in trustworthiness compared to models originating from academic research (\eg, CogVideo~\cite{hong2022cogvideo} and Show-1~\cite{zhang2023show}). A possible explanation is that industrial organizations might have more diverse in-house data and emphasize internal reviews focused on trustworthiness during model development. We encourage users and researchers to exercise caution when using video generation models, keeping safety and ethical considerations in mind.
}

\section{Conclusion}

With the growing focus on video generation, comprehensive evaluation of these models is essential to assess current advancements and guide future research.
In this work, we take the first step forward and propose \textbf{\textit{VBench}}, a comprehensive benchmark suite for evaluating video generation models. 
\extension{On top of VBench, we propose \textbf{\textit{VBench++}} to support image-to-video evaluation, and trustworthiness evaluation. We also continually include newly released models to our leaderboard to drive forward the field of video generation.
With its \textit{multi-dimensional, human-aligned, insight-rich, and versatile} properties, VBench++ could play vital roles for evaluating future video generation models and inspiring further advancements in video generation.
We believe that VBench++ is a significant contribution to the video generation and evaluation community.
}

\smallTitle{Acknowledgement.} 
This study is supported by the Ministry of Education, Singapore, under its MOE AcRF Tier 2 (MOET2EP20221-0012), NTU NAP, and under the RIE2020 Industry Alignment Fund - Industry Collaboration Projects (IAF-ICP) Funding Initiative, as well as cash and in-kind contribution from the industry partner(s); the National Key R\&D Program of China (2022ZD0160201); National Natural Science Foundation of China (62076119, 61921006, 62102150); the Science and Technology Commission of Shanghai Municipality (23YF1461900, 23QD1400800); and Shanghai Artificial lntelligence Laboratory.

{
    \small
    \bibliographystyle{IEEEtran}
    \bibliography{main}
}

\end{document}